\documentclass{jingdong}

\usepackage[utf8]{inputenc}
\usepackage{amsmath,amssymb}
\usepackage{enumitem}
\usepackage{float}
\usepackage[ruled,linesnumbered,lined,shortend]{algorithm2e}
\SetAlFnt{\small}
\SetAlCapFnt{\small}
\SetAlCapNameFnt{\small}
\SetAlgoNlRelativeSize{-1}
\SetNlSkip{0.5em}
\newcommand{\method}{AdCoEvo}

\title{\method: A Co-Evolving Multi-Agent Advertising Delivery System}
\renewcommand{\authorfont}{\fontsize{11}{14}\selectfont}
\author[1,2,*,\ddagger]{Tianzhu~Hu}
\author[3,*,\ddagger]{Hang~Yu}
\author[4]{Zeyuan~Zhao}
\author[4]{Zeyu~Pan}
\author[4]{Huan~Luo}
\author[4,\dagger]{Chun~Gan}
\author[4]{Mingzhe~Cao}
\author[4]{Liubin~Wang}
\author[1,\dagger]{Mingming~Li}
\author[1]{Jiao~Dai}
\author[1]{Jizhong~Han}
\author[1,2]{Songlin~Hu}

\newcommand{\AdCoEvoAuthorEmails}{%
  {\affiliationfont
  \begin{minipage}{\dimexpr\textwidth-1pt\relax}
  \centering
  \email{hutianzhu@iie.ac.cn}, \email{hayu0853@uni.sydney.edu.au}\\
  \href{mailto:zhaozeyuan.6@jd.com,panzeyu3@jd.com,luohuan.12@jd.com,ganchun1@jd.com,caomingzhe1@jd.com,wangliubin.ubi@jd.com}{\texttt{\{zhaozeyuan.6, panzeyu3, luohuan.12, ganchun1, caomingzhe1, wangliubin.ubi\}@jd.com}}\\
  \href{mailto:limingming@iie.ac.cn,daijiao@iie.ac.cn,hanjizhong@iie.ac.cn,husonglin@iie.ac.cn}{\texttt{\{limingming, daijiao, hanjizhong, husonglin\}@iie.ac.cn}}
  \end{minipage}}%
}

\renewcommand{\affiliationlist}{%
  \affiliationformat[1]{Institute of Information Engineering, Chinese Academy of Sciences}\\
  \affiliationformat[2]{School of Cyberspace Security, University of Chinese Academy of Sciences}\\
  \affiliationformat[3]{The University of Sydney}\quad
  \affiliationformat[4]{JD.COM}\par
  \vspace{2mm}
  \AdCoEvoAuthorEmails
}

\contribution[*]{Equal contribution}
\contribution[\dagger]{Corresponding authors}
\contribution[\ddagger]{Work was done during internship.}

\abstract{%
Digital advertising is an important means for platforms to reach potential users and drive business growth. Advertising delivery is the core link between marketing objectives and actual returns. Existing methods have achieved partial automation in individual tasks such as bidding and budget allocation, but they still rely on human operators for overall coordination, making it difficult to support large-scale operations. They also often fail to achieve optimal performance, and individual experience remains difficult to turn into reusable operational capabilities. In addition, as advertising markets change, advertising delivery systems need timely strategy updates to remain effective in dynamic markets. To automate the entire advertising operations process and improve adaptability to market changes, we propose AdCoEvo, a co-evolving multi-agent advertising delivery system. Specifically, the system connects delivery decisions across different timescales and operational granularities through role specialization. Diagnostic results guide account planning and strategy adjustment in the next period. The multi-agent system updates its model parameters based on actual delivery outcomes and continually improves delivery strategies through mutual feedback among planning, adjustment, and diagnosis, thereby achieving co-evolution. Experiments show that, relative to the rule-based operational baseline (Rule), AdCoEvo increases advertiser value by 14.02\% while reducing cost per acquisition by 7.07\%. Compared with a multi-agent system with fixed parameters, AdCoEvo increases advertiser value by 3.96\% and reduces cost per acquisition by 1.24\%, indicating that co-evolution can yield greater delivery returns.
}
\checkdata[Code]{\url{https://github.com/jd-opensource/AdCoEvo}}

\begin{document}

\maketitle

\section{Introduction}
\label{sec:1}

\begin{figure}[!t]
\centering
\includegraphics[width=\linewidth]{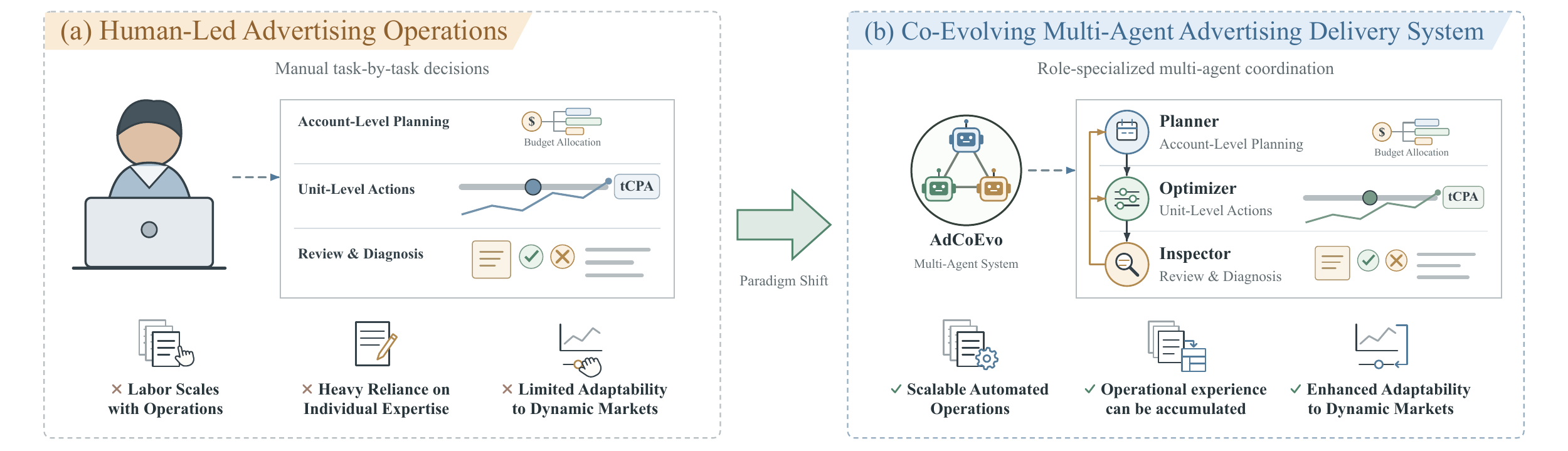}
\caption{From manual to multi-agent advertising delivery.}
\label{fig:intro-paradigm}
\end{figure}

Digital advertising is an important channel for platforms to reach potential users and drive business growth. Turning marketing objectives into returns requires continuous high-quality advertising delivery decisions under budget constraints~\cite{zhang2014ortb}. As shown on the left of Figure~\ref{fig:intro-paradigm}, existing delivery workflows still rely heavily on manual configuration and experience-based judgment by operators~\cite{sun2026adpilot}. As operations grow, limited manpower and responsiveness make this approach difficult to scale and often prevent optimal performance~\cite{sun2026adpilot}. Operators' accumulated experience is also hard to turn into reusable operational capabilities for later delivery.

Prior work has automated individual tasks, such as bidding~\cite{mou2022sorl,zhang2014ortb} and budget allocation~\cite{wang2024hibid,nuara2018adcomb,song2026dara}, but overall advertising operations still rely on human coordination. More recently, large language model agents have begun to take on decisions across the entire advertising delivery process. AdPilot uses a Reflector--Actor workflow: it first generates a diagnosis from current metrics and historical trends, then makes delivery decisions based on that diagnosis, and improves agent capabilities through a series of post-training steps~\cite{sun2026adpilot}.

Training policies for the entire delivery process requires linking business feedback to specific decisions. Budget planning and ad-unit adjustments continue to affect subsequent delivery, and their effects need to be evaluated through cumulative returns; post-delivery diagnosis takes effect through planning and adjustment in the next period. Organizing multi-agent collaboration by operational responsibility therefore also requires constructing training signals according to each decision's scope of influence and feedback timing, and scheduling role updates after feedback settlement, enabling models to co-evolve during continuous delivery.

Meanwhile, the advertising delivery market is not static: a continually changing auction environment may create a mismatch between historical strategies and current conditions, requiring timely updates to delivery strategies~\cite{mou2022sorl,croissant2020realtime}. Beyond autonomous execution, we therefore need to study how multi-agent systems use continuous business feedback to adjust strategies and jointly adapt to market changes through collaboration.

To automate the entire advertising delivery process and improve adaptability to market changes, we propose AdCoEvo, a co-evolving multi-agent advertising delivery system. As shown on the right of Figure~\ref{fig:intro-paradigm}, the system connects decisions across different timescales and operational granularities by assigning separate responsibilities for account-level budget planning, ad-unit delivery strategy adjustment, and end-of-period diagnosis. Diagnostic results guide account planning and strategy adjustment in the next period. After offline training, the system updates one role per period in rotation using settled feedback, while keeping other roles' parameters fixed. The updated policy takes effect in the next period. Through joint execution, it changes the delivery states and training feedback encountered by other roles, which adapt their policies in later rotations. We call this mutual adaptation across periods co-evolution.
Experiments show that AdCoEvo improves advertiser value (ADV) and reduces cost per acquisition (CPA) in both environments.\footnote{\raggedright $\mathrm{ADV}=\sum(\mathrm{tCPA}\times\mathrm{pCVR})$ over valid impressions~\cite{badanidiyuru2026monotonicity}; \mbox{$\mathrm{CPA}=\text{total spend}/\text{total conversions}$}~\cite{googleads_average_cpa}. Here, tCPA is the target cost per acquisition and pCVR is the predicted conversion probability.} Compared with a multi-agent system with fixed parameters, co-evolution can yield greater delivery returns.

Our main contributions are as follows:

\begin{enumerate}
\item We propose AdCoEvo, a co-evolving multi-agent advertising delivery system that connects account budget planning, ad-unit delivery strategy adjustment, and post-delivery diagnosis through role specialization across different timescales and operational granularities, enabling autonomous delivery across periods.
\item We develop a co-evolution framework for continuous advertising delivery in which the multi-agent system updates policies using settled business feedback and jointly adapts to changes in delivery as its decisions influence one another.
\begin{samepage}
\item We validate the system's delivery performance across different market sequences and compare role compositions, training stages, and role-update schemes to analyze gains from role collaboration, operational capability development, and co-evolution.\par
\end{samepage}
\end{enumerate}

\section{Related Work}
\label{sec:2}

\subsection{Advertising Delivery}
\label{sec:2-1}

Research on advertising delivery has largely focused on automating individual tasks. Automated bidding has progressed from optimizing bid functions under budget constraints~\cite{zhang2014ortb} to using reinforcement learning~\cite{wu2018drlb,cai2017rlb} or generative modeling~\cite{guo2024aigb} for sequential decision-making. RTBAgent incorporates expert knowledge and reflection on past experience into bidding decisions~\cite{cai2025rtbagent}. AdComB jointly optimizes budget allocation and bidding~\cite{nuara2018adcomb}, while HiBid uses a hierarchical mechanism to connect cross-channel budgets with within-channel bidding~\cite{wang2024hibid}. AuctionNet evaluates automated bidding through traffic generation and auction interactions~\cite{auctionnet2024}.

AdPilot supports end-to-end autonomous advertising delivery through a Reflector--Actor workflow informed by current states and historical memory. Post-training further improves its delivery decisions~\cite{sun2026adpilot}. AdCoEvo organizes operational responsibilities into roles with separate policies. Period-end diagnosis guides account planning and strategy adjustment in the next period, while settled feedback supports role-policy updates during continuous delivery.

\subsection{Multi-Agent Collaboration}
\label{sec:2-2}

The core advantage of multi-agent systems is decomposing complex processes into roles with clear responsibilities. Roles address distinct objectives and contexts, then make collective decisions through interaction. Studies have applied complementary roles and task decomposition to complex problem solving and workflow automation, supporting multi-step tasks that are difficult for a single agent to complete independently~\cite{li2023camel,wu2024autogen,hong2024metagpt,qian2024chatdev,tao2024magis}.

Current research on multi-agent systems organizes role collaboration through standardized workflows, dynamic task allocation, and runtime supervision~\cite{hong2024metagpt,niu2025flow,zhou2026mass,lin2026supervisoragent,chen2024agentverse}. It also trains models using joint execution trajectories, role evaluations, or interaction rewards to improve collaborative decision-making capabilities~\cite{motwani2025malt,xue2026comas}. Our work focuses on the collaboration and continual evolution of multi-agent systems in advertising delivery tasks.

\subsection{Agentic RL}
\label{sec:2-3}

Agentic RL uses feedback from environmental interactions to optimize multi-step decision-making and tool-use policies~\cite{jin2025searchr1,du2026codegym,qi2025webrl}. For example, ReTool uses tool-call outcomes to improve when and how tools are called~\cite{feng2026retool}, while AgentFlow uses multi-turn trajectories to improve planning policies~\cite{li2026agentflow}. MARTI uses multi-agent interaction trajectories for distributed training updates~\cite{zhang2026marti}.

We continually update role policies through alternating reinforcement learning, focusing on mutual adaptation among roles during continuous delivery. Training signals reflect each role's scope of influence and feedback timing. Cross-period diagnosis and subsequent joint execution allow each role's policy updates to shape feedback for other roles' subsequent decisions and learning.

\section{Problem Formulation}
\label{sec:3}

We formulate continuous advertising delivery as sequential decision-making, with objectives for value generation and conversion costs under budget and adjustment-count constraints.

\subsection{Modeling Delivery Decisions}
\label{sec:3-1}

During delivery, the system plans ad-unit budgets within each account and adjusts control tCPA based on the current state. The fixed underlying bidder uses these settings to bid for individual traffic opportunities. At the end of the period, the system generates diagnostic guidance from settled outcomes to inform account planning and strategy adjustment in the next period.

Index decisions chronologically under system policy $\pi$. Decision $i$ uses currently observable information:
\begin{equation}
u_i\sim\pi(\cdot\mid h_i).
\label{eq:decision-process}
\end{equation}

\begin{itemize}
\item \textbf{Decision information $h_i$} consists of the current decision type and its observable delivery state, historical execution records, and settled feedback. Historical diagnostic guidance forms part of this information and informs subsequent account planning and strategy adjustment.
\item \textbf{Decision output $u_i$}: an account's ad-unit budget allocations, control tCPA, and diagnostic guidance.
\end{itemize}

\subsection{Business Objectives and Constraints}
\label{sec:3-2}

The system measures the business outcomes of delivery decisions using ADV and CPA, which reflect value generation and conversion costs, respectively. Let $\mathcal A$ denote the set of controlled accounts and $\mathcal U_a$ the set of ad units in account $a$. The total ADV across all controlled accounts in period $e$ is
\begin{equation}
F_e=\sum_{a\in\mathcal A}\sum_{j\in\mathcal U_a}d_{a,j}q_{e,a,j},
\label{eq:advertiser-value}
\end{equation}

Here, $d_{a,j}$ is the environment-provided initial tCPA of the ad unit, and $q_{e,a,j}$ is the sum of predicted conversion probabilities for its valid impressions within the period. ADV is always calculated using the initial tCPA. Control tCPA changes affect ADV through delivery outcomes, while its valuation basis stays fixed.

Let $\bar F_e=F_e/|\mathcal A|$ denote average ADV per account, and $C_e,N_e$ denote total spend and conversions across controlled accounts in period $e$. Aggregate CPA is $C_e/N_e$. We optimize these two objectives subject to budget and adjustment-count constraints:
\begin{equation}
\begin{aligned}
\underset{\pi\in\Pi}{\operatorname{maximize}}\quad
&\left(
\mathbb E_\pi[\bar F_e],
-\frac{\mathbb E_\pi[C_e]}{\mathbb E_\pi[N_e]}
\right),\\
\text{subject to}\quad
&\sum_{j\in\mathcal U_a}b_{e,a,j}=B_{e,a},
&&\forall e,a,\\
&b_{e,a,j}\geq0,\qquad v_{e,a,j}\leq K,
&&\forall e,a,j.
\end{aligned}
\label{eq:delivery-objective}
\end{equation}

\begin{itemize}
\item \textbf{Policy set $\Pi$}: policies use only information observable at decision time; CPA requires $\mathbb E_\pi[N_e]>0$.
\item \textbf{Budget constraint:} $B_{e,a}$ is the budget of account $a$ in period $e$, and $b_{e,a,j}$ is the total budget limit of ad unit $j$. Reallocation among ad units preserves the account's total budget.
\item \textbf{Adjustment-count constraint:} $v_{e,a,j}$ counts valid tCPA adjustments executed for ad unit $j$ in period $e$; $K$ is the maximum count per ad unit per period.
\end{itemize}

\section{Method}
\label{sec:4}

\begin{figure}[!t]
\centering
\includegraphics[width=\linewidth]{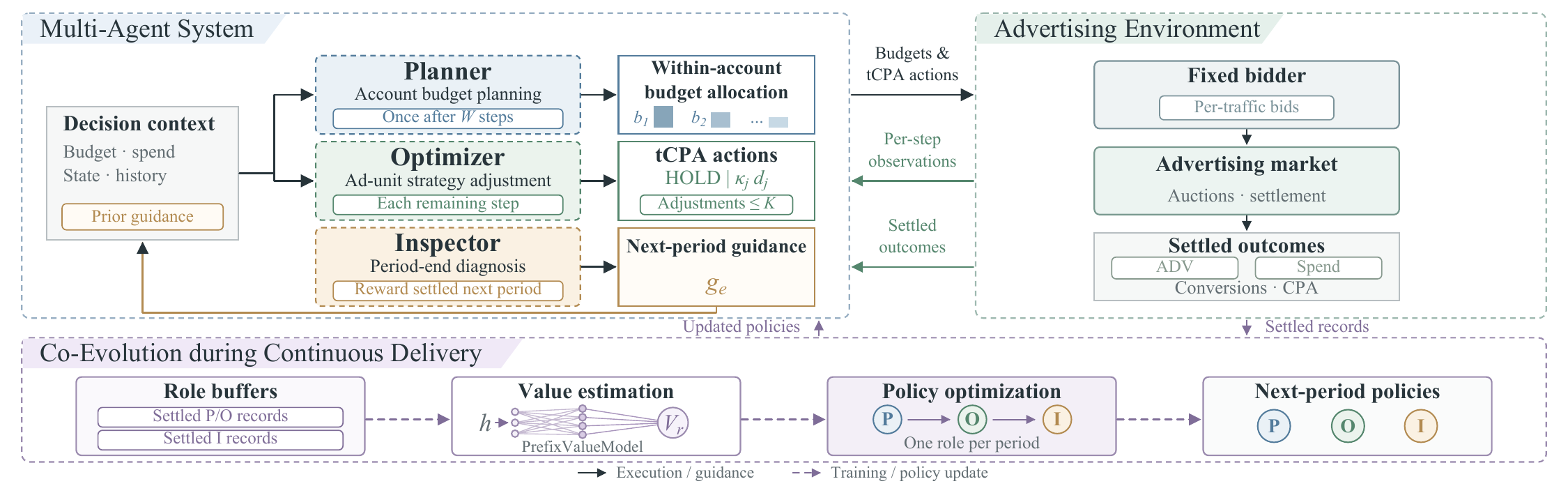}
\caption{Role collaboration and continuous co-evolution in AdCoEvo. The upper panel shows delivery execution, and the lower panel shows role-policy updates.}
\label{fig:adcoevo-overview}
\end{figure}

AdCoEvo organizes advertising delivery through role collaboration. As shown in the upper part of Figure~\ref{fig:adcoevo-overview}, Planner handles account budget planning, Optimizer adjusts ad-unit delivery strategies, and Inspector turns settled period outcomes into diagnostic guidance for the next period. Together, planning, adjustment, and diagnosis form a collaborative process across periods.

The system adopts two-stage training. The first stage develops operational decision-making capabilities through supervised fine-tuning and offline reinforcement learning. The second stage, shown in the lower part of Figure~\ref{fig:adcoevo-overview}, updates role parameters using settled business feedback during continuous delivery, enabling co-evolution. The offline reinforcement learning and co-evolution algorithms are given in Appendices~\ref{sec:offline-algorithm} and~\ref{sec:coevo-algorithm}, respectively.

\subsection{Role Specialization and Collaborative Execution}
\label{sec:4-1}

Account budget planning, ad-unit delivery strategy adjustment, and end-of-period diagnosis differ in their scope of influence and feedback timing. AdCoEvo accordingly assigns these responsibilities to Planner, Optimizer, and Inspector and maintains a separate policy $\pi_{\theta_r}$ for each role, where $r\in\{P,O,I\}$ and $\theta_r$ denotes the role's model parameters. This organization associates execution records for different responsibilities with their corresponding training feedback. Planner and Optimizer make decisions for one account and its ad units at a time, whereas Inspector aggregates period outcomes across all controlled accounts.

After observing $W$ time steps, Planner uses observed delivery performance to allocate the account's remaining budget, outputting a proportion $\alpha_j$ for each ad unit. Unit $j$ then has total budget limit
\begin{equation}
b_j=c_j^{\mathrm{pre}}+R\alpha_j,
\label{eq:budget-allocation}
\end{equation}

Here, $c_j^{\mathrm{pre}}$ is the cumulative spend of ad unit $j$ before planning, $R=B-\sum_{j\in\mathcal U}c_j^{\mathrm{pre}}$ is the account's remaining budget, and $B$ and $\mathcal U$ denote the account's total budget and set of ad units, respectively. The allocation proportions satisfy $\alpha_j\geq0$ and $\sum_{j\in\mathcal U}\alpha_j=1$.

After budget planning, Optimizer determines at every remaining time step whether each ad unit in the account needs a delivery strategy adjustment. Ad-unit actions include HOLD and a multiplier adjustment relative to the initial tCPA. If an adjustment is selected and the unit's number of valid adjustments in the current period has not reached $K$, Optimizer selects $\kappa_j$ from the available multiplier set $\mathcal K$ and updates the control tCPA to
\begin{equation}
\widehat d_j=\kappa_jd_j,
\label{eq:tcpa-adjustment}
\end{equation}

Here, $d_j$ is the environment-provided initial tCPA. The current control value is retained for HOLD or when valid adjustments reach $K$.

After period settlement, Inspector aggregates delivery performance, budget utilization, and action records across the controlled accounts to generate diagnostic guidance $g_e$ as part of the decision inputs of Planner and Optimizer in the next period.

\subsection{Role Reward Construction}
\label{sec:4-2}

Training feedback for Planner and Optimizer accounts for both value generation and cost control. For ADV, spend, and conversions $F,C,N$ within the corresponding aggregation scope, we define the proxy return
\begin{equation}
Q(F,C,N)
=
\frac{F-\eta(C-\rho N)}{s},
\label{eq:coevo-reward-score}
\end{equation}

Here, $\rho$ is the mean initial tCPA of ad units in the settlement window, weighted by conversion counts; $\eta$ weights the cost deviation $C-\rho N$; and $s$ scales the reward.

Role rewards reflect the scope of influence and feedback timing of each decision. Planner coordinates the account's period budget and evaluates resource allocation using outcomes from observation end to period end. Optimizer continually adjusts ad-unit delivery strategies, computes stepwise returns from value generation and the cost deviation, and accumulates subsequent returns during training to account for the lasting effects of its actions. Both rewards combine the current account's return with a weighted overall return. Inspector provides diagnostic guidance to Planner and Optimizer in the next period. Its reward is computed from changes in overall feedback between adjacent periods once the next period's delivery outcomes are settled:
\begin{equation}
r_a^P=Q_a+\omega Q_{\mathrm{all}},\qquad r_{a,t}^O=Q_{a,t}+\omega Q_{\mathrm{all},t},\qquad r_{e-1}^I=\mathbf w^\top(\mathbf y_e-\mathbf y_{e-1}).
\label{eq:role-rewards}
\end{equation}

Here, $Q_a,Q_{\mathrm{all}}$ aggregate outcomes from observation end to period end for the current account and all controlled accounts, respectively; time-indexed terms use current-step increments. The parameter $\omega$ weights the overall reward. The period feedback vector $\mathbf y_e$ combines standardized ADV, conversions, the CPA constraint margin, and the forced-HOLD rate across all controlled accounts, with fixed weights $\mathbf w$. Unsettled diagnoses are excluded from training. Calculations appear in Appendices~\ref{sec:reward-po}--\ref{sec:reward-inspector}.

\subsection{Offline Development of Operational Capabilities}
\label{sec:4-3}

We first apply supervised fine-tuning~\cite{ouyang2022instructgpt,zeng2024agenttuning} on role-specific demonstration trajectories to learn action semantics and structured output specifications. Offline reinforcement learning then uses delivery outcomes as training feedback aligned with each role's responsibilities to improve operational decisions further.

To compare different decisions from the same delivery state, the system fixes the other roles' policies, samples $G$ candidate responses for the target role, and executes them separately, with $J_g^r$ denoting the return of candidate $g$. Planner and Optimizer construct leave-one-out advantages~\cite{ahmadian2024rloo}; Inspector evaluates the delivery improvement from guidance against the baseline return $J_0$ of a branch in which neither Planner nor Optimizer receives diagnostic guidance. The candidate advantages are
\begin{equation}
A_g^r=
\begin{cases}
J_g^r-\dfrac{1}{G-1}\displaystyle\sum_{k\ne g}J_k^r,
&r\in\{P,O\},\\[6pt]
J_g^I-J_0,
&r=I.
\end{cases}
\label{eq:offline-advantage}
\end{equation}

Planner and Optimizer advantages are normalized by a scale, while Inspector retains the return difference relative to the baseline without diagnostic guidance. We denote them by $\widehat A_g^r$. Using the sequence-level clipping approach of GSPO~\cite{zheng2025gspo}, we apply decision advantages to policy optimization over entire responses. For role $r$ with valid candidates $\mathcal I_r$, we optimize
\begin{equation}
\mathcal L_{\mathrm{off}}^r(\theta_r)
=
-\frac{1}{|\mathcal I_r|}
\sum_{g\in\mathcal I_r}
\min\!\left[
\varrho_g\widehat A_g^r,\,
\operatorname{clip}
\!\left(\varrho_g,1-\varepsilon^-,1+\varepsilon^+\right)
\widehat A_g^r
\right]
+\beta\widehat D_{\mathrm{KL}}.
\label{eq:offline-loss}
\end{equation}

Here, $\varrho_g$ is the length-normalized sequence importance ratio of the current policy to the sampling policy, and $\widehat D_{\mathrm{KL}}$ measures the current policy's deviation from a fixed reference policy. The parameters $\varepsilon^-,\varepsilon^+$ and $\beta$ control the clipping range and KL penalty strength, respectively. The calculations for advantage processing, the sequence ratio, and the KL term are detailed in Appendix~\ref{sec:offline-optimization}.

\subsection{Co-Evolution during Continuous Delivery}
\label{sec:4-4}

Offline training compares candidate decisions while keeping the other roles' policies fixed. During continuous delivery, updated roles participate in joint execution again, changing subsequent delivery states and business feedback, to which other roles must also adapt their decisions. AdCoEvo draws on SAO's single-trajectory asynchronous update idea~\cite{hou2026sao} to construct training signals from settled execution trajectories.

At period end, the system settles current P/O rewards and constructs period feedback $\mathbf y_e$. Inspector generates diagnostic guidance $g_e$ for the next period from the delivery outcomes. The system then settles the reward for the previous diagnosis $g_{e-1}$, adds settled records to role buffers, and updates one role in rotation $(P,O,I)$. Other role parameters remain fixed, and the new policy takes effect in the next period.

\paragraph*{Value Estimation and Policy Optimization.}

During continuous delivery, the system executes one response at each decision point. Each role therefore maintains a separate lightweight value model, PrefixValueModel $V_{\phi_r}(h)$, which predicts a scalar return from the input before response generation. The model is continually trained on settled feedback to provide a baseline for policy updates.

Let $V_i^-$ denote the value prediction for node $i$ before the current round of updates. Planner and Inspector subtract this baseline from their settled rewards. Optimizer combines current-step rewards and subsequent value estimates through GAE~\cite{schulman2016gae}:
\begin{equation}
\widehat A_i^r=
\begin{cases}
r_i^r-V_i^-,
&r\in\{P,I\},\\[4pt]
\operatorname{GAE}_i(\{r_k^O,V_k^-\}),
&r=O.
\end{cases}
\label{eq:coevo-advantage}
\end{equation}

Optimizer successor nodes follow decision time within the same account and period.

Advantages remain fixed during the current round of policy optimization. The value model is trained independently by regression, without propagating its gradients into the role language model. The entire response corresponding to a decision shares the same action advantage. For a response $z$ of length $L$, input $h$, and action advantage $\widehat A$, we optimize
\begin{equation}
\mathcal L_{\mathrm{co-evo}}(\theta_r)
=
-\frac{\widehat A}{Z}
\sum_{\ell=1}^{L}
m_\ell\bar\varrho_\ell
\log\pi_{\theta_r}(z_\ell\mid h,z_{<\ell}),
\label{eq:coevo-loss}
\end{equation}

Here, $m_\ell$ is a token-selection mask based on the probability ratio between the current and behavior policies, $\bar\varrho_\ell$ is a clipped importance weight with its gradient stopped, and $Z=\sum_\ell m_\ell>0$ is the number of valid tokens. KL is used only for monitoring and triggering rollback. Implementation details of advantage estimation, value regression, and update control are provided in Appendices~\ref{sec:gae-details}--\ref{sec:sao-details}.

Through subsequent joint execution, updated policies change the delivery states and feedback encountered by other roles. These roles adjust their policies accordingly in later rotations, enabling the system to continually evolve through collaboration.

\subsection{Inference Time Complexity Analysis}
\label{sec:4-5}

AdCoEvo's per-period inference overhead depends on both the number of model calls and the cost of each call. Let $n$ denote the number of decision points per period and $|\mathcal A|$ the number of controlled accounts. Planner is called once per account after observation, Optimizer is called once per account at each decision point after observation, and Inspector is called once globally at the end of the period. The number of model calls per period is therefore $O(|\mathcal A|n)$.

Let $L$ denote the combined input and output length per call, which is affected by the number of ad units per account and the length of historical records. For a fixed model size with KV caching, the computational cost of Transformer self-attention is $O(L^2)$ per call, yielding the following per-period inference complexity
\begin{equation}
\mathcal C_{\mathrm{infer}}
=
O\!\left(|\mathcal A|nL^2\right).
\label{eq:inference-complexity}
\end{equation}

\section{Experiments}
\label{sec:5}

We compare AdCoEvo with operational baselines and foundation models in delivery performance and overhead across continuous market sequences. Ablations examine the roles of collaboration, operational capability development, and diagnostic guidance. Role-update comparisons and market-switch experiments further assess whether co-evolution continually improves delivery, while budget and parameter experiments test whether these gains persist under different settings.

\FloatBarrier

\subsection{Experimental Setup}
\label{sec:5-1}

\textbf{Experimental environment:} Experiments use the AuctionNet simulator~\cite{auctionnet2024} with its official period length and traffic volume. The environment contains 48 ad units that can purchase traffic. We control 12 units organized into 3 accounts, each containing 4 units, all using Abid as the fixed underlying bidder. We use two environments, AuctionNet (Sparse) and AuctionNet (General). The former uses the public simulation environment, where each controlled account's initial budget is $1/12$ of the total budget of all units. For AuctionNet (General), we calibrate the pValue generation scale, ad-unit budgets, and CPA constraints using the public General dataset, while keeping the other auction mechanisms unchanged. Detailed configurations are provided in Appendix~\ref{sec:environment-config}.

\textbf{Evaluation metrics:} Delivery performance is measured using ADV, CPA, and a composite score. For ADV, which uses tCPA-weighted predicted conversion values~\cite{badanidiyuru2026monotonicity}, we report the per-period average of total ADV across all controlled accounts within the evaluation scope. CPA is total spend divided by total conversions over the same evaluation scope~\cite{googleads_average_cpa}. The invalid output rate $\mathrm{IR}$ is the fraction of responses that violate the required format or action specifications, such as parsing failures, incorrect ad-unit decision counts, or tCPA multipliers outside the allowed set. Each invalid response is counted once. With No Operation under the same environment and budget as the reference, method $m$'s composite score is
\begin{equation}
\mathrm{Score}_m
=\frac{\mathrm{ADV}_m}{\mathrm{ADV}_0}
\cdot\frac{\mathrm{CPA}_0}{\mathrm{CPA}_m}
-\mathrm{IR}_m.
\label{eq:score}
\end{equation}

Here, subscript $0$ denotes No Operation, whose score is 1. Invalid output rates are reported as percentages in tables and used as decimal fractions in the formula.

\begin{samepage}
\textbf{Comparison methods:} We use No Operation, Rule, and AdPilot\textsuperscript{*} as advertising delivery baselines to compare delivery performance across different approaches to operational decision-making. No Operation makes no operational adjustments, while Rule adjusts budgets and tCPA using hand-crafted CPA-based rules. AdPilot\textsuperscript{*} follows AdPilot's Reflector--Actor workflow~\cite{sun2026adpilot}, where a single agent generates actions informed by delivery diagnosis. AdCoEvo (Fixed) retains all roles and the diagnostic workflow, with model parameters fixed at the offline GSPO checkpoint during evaluation. We also evaluate representative closed- and open-source foundation models~\cite{deepseekai2026deepseekv4,kimiteam2026kimik3,zai2026glm52,qwen2026qwen35}\footnote{Closed-source model documentation: \href{https://developers.openai.com/api/docs/models/gpt-5.6-terra}{GPT-5.6-Terra}, \href{https://www.anthropic.com/news/claude-sonnet-5}{Claude-Sonnet-5}, and \href{https://deepmind.google/models/model-cards/gemini-3-1-flash-lite/}{Gemini-3.1-Flash-Lite}.} that directly perform advertising delivery tasks, comparing delivery performance, output validity, and operational overhead.
\end{samepage}

\textbf{Implementation and evaluation:} AdCoEvo uses Qwen3-8B~\cite{yang2025qwen3} with separate LoRA parameters~\cite{hu2022lora} for each role. AdPilot\textsuperscript{*} uses the same backbone and follows AdPilot's post-training procedure~\cite{sun2026adpilot}, with SFT followed by GRPO. Its model parameters remain fixed during formal evaluation. We set the observation length to $W=12$, with at most $K=8$ tCPA adjustments per ad unit per period.

Each experimental group uses five independent market sequences of 50 consecutive delivery periods, matched across methods. These sequences were not used for SFT, offline reinforcement learning, or selecting the default parameters. AdCoEvo independently starts each sequence from the same offline GSPO checkpoint. AuctionNet regenerates market conditions, including traffic and predicted conversion probabilities, each period, while role policies and diagnostic guidance carry over within the sequence.

\FloatBarrier

\subsection{Overall Delivery Performance}
\label{sec:5-2}

We first examine whether AdCoEvo can achieve better delivery outcomes than fixed operational rules derived from human experience. As shown in Table~\ref{tab:overall}, relative to Rule, AdCoEvo increases ADV by 14.02\% and reduces CPA by 7.07\% in the Sparse environment, while also improving both metrics in the General environment. Human-designed rules can trigger adjustments based on delivery states, but their decision logic is specified in advance. AdCoEvo instead uses delivery feedback for policy learning and continuous updates, allowing the models to accumulate experience throughout delivery.

\begin{table}[!t]
\centering
\caption{Delivery performance across market sequences. Overall aggregates results from five 50-period sequences.}
\label{tab:overall}
\begingroup
\fontsize{7.5}{9.0}\selectfont
\setlength{\tabcolsep}{1.2pt}
\begin{tabular*}{\linewidth}{@{\extracolsep{\fill}}l*{10}{r}@{\hspace{5pt}}rrr@{}}
\toprule
Method & \multicolumn{2}{c}{Sequence 1} & \multicolumn{2}{c}{Sequence 2} & \multicolumn{2}{c}{Sequence 3} & \multicolumn{2}{c}{Sequence 4} & \multicolumn{2}{c}{Sequence 5} & \multicolumn{3}{c}{Overall} \\
\cmidrule(lr){2-3}\cmidrule(lr){4-5}\cmidrule(lr){6-7}\cmidrule(lr){8-9}\cmidrule(lr){10-11}\cmidrule(l){12-14}
 & ADV $\uparrow$ & CPA $\downarrow$ & ADV $\uparrow$ & CPA $\downarrow$ & ADV $\uparrow$ & CPA $\downarrow$ & ADV $\uparrow$ & CPA $\downarrow$ & ADV $\uparrow$ & CPA $\downarrow$ & ADV $\uparrow$ & CPA $\downarrow$ & Score $\uparrow$ \\
\midrule
\multicolumn{14}{c}{\rule[-0.5ex]{0pt}{2.6ex}\textit{AuctionNet (Sparse)}} \\
No Operation & 26,440.6 & 115.575 & 26,372.9 & 115.676 & 26,175.7 & 117.539 & 26,598.1 & 116.739 & 26,286.4 & 115.531 & 26,374.7 & 116.207 & 1.0000 \\
Rule & 28,680.3 & 104.124 & 28,333.9 & 103.629 & 28,376.6 & 105.553 & 28,537.4 & 104.623 & 28,195.3 & 103.658 & 28,424.7 & 104.315 & 1.2006 \\
AdPilot\textsuperscript{*} & 28,090.3 & 118.479 & 27,012.3 & 118.847 & 27,103.3 & 120.397 & 27,046.0 & 119.133 & 26,876.5 & 118.358 & 27,225.7 & 119.040 & 1.0077 \\
AdCoEvo (Fixed) & 31,720.6 & 98.368 & 30,709.7 & 98.583 & 30,615.4 & 100.112 & 31,675.0 & 99.104 & 31,154.9 & 97.600 & 31,175.1 & 98.150 & 1.3995 \\
\textbf{AdCoEvo} & \textbf{32,125.4} & \textbf{96.807} & \textbf{32,350.1} & \textbf{96.217} & \textbf{32,725.4} & \textbf{98.087} & \textbf{32,341.5} & \textbf{98.220} & \textbf{32,500.9} & \textbf{96.361} & \textbf{32,408.7} & \textbf{96.936} & \textbf{1.4731} \\
\noalign{\vskip0.2ex
  \hbox to\linewidth{\leaders\hbox{\kern1.5pt\vrule width3pt height0.3pt\kern1.5pt}\hfill}
  \vskip0.2ex}
\multicolumn{14}{c}{\rule[-0.5ex]{0pt}{2.6ex}\textit{AuctionNet (General)}} \\
No Operation & 27,612.0 & 8.864 & 28,094.9 & 8.685 & 27,782.8 & 8.875 & 28,306.5 & 8.738 & 28,203.9 & 8.677 & 28,000.0 & 8.767 & 1.0000 \\
Rule & 29,917.9 & 7.199 & 30,282.6 & 7.084 & 30,244.9 & 7.266 & 30,513.2 & 7.143 & 30,147.7 & 7.077 & 30,221.3 & 7.153 & 1.3229 \\
AdPilot\textsuperscript{*} & 28,688.2 & 9.122 & 28,604.6 & 8.879 & 28,223.2 & 9.147 & 28,762.7 & 8.979 & 28,336.0 & 8.806 & 28,522.9 & 8.985 & 0.9940 \\
AdCoEvo (Fixed) & 31,965.4 & 7.112 & 31,375.8 & 7.025 & 30,914.1 & 7.139 & 32,294.7 & 7.258 & 31,541.2 & 7.203 & 31,618.2 & 7.147 & 1.3852 \\
\textbf{AdCoEvo} & \textbf{33,164.4} & \textbf{7.069} & \textbf{33,009.5} & \textbf{6.892} & \textbf{32,507.9} & \textbf{7.068} & \textbf{33,744.4} & \textbf{6.944} & \textbf{32,334.3} & \textbf{6.975} & \textbf{32,952.1} & \textbf{6.989} & \textbf{1.4763} \\
\bottomrule
\end{tabular*}
\endgroup

\end{table}

The comparison with AdPilot\textsuperscript{*} further examines how end-to-end delivery decisions are organized. Both use the same backbone model. With both systems' parameters fixed during evaluation, AdCoEvo (Fixed) achieves 14.51\% higher ADV and 17.55\% lower CPA in Sparse. Compared with a single agent that handles delivery decisions, AdCoEvo assigns operational responsibilities to roles that learn separately and provide feedback to one another, enabling them to collaborate and co-evolve toward a shared business objective.

Across all five market sequences in both environments, AdCoEvo achieves the highest ADV and lowest CPA among the compared methods. These results indicate its adaptability to changing markets.

\subsection{Comparison with Foundation Models}
\label{sec:5-3}

We compare with foundation models to assess advertising-specific operational capabilities. As shown in Table~\ref{tab:general-models}, AdCoEvo achieves the highest ADV, lowest CPA, and highest composite score. Compared with GPT-5.6-Terra, the best-performing foundation model overall, it increases ADV by 3.73\% and reduces CPA by 3.23\%. With a smaller model, domain-specific training and co-evolution can build operational capabilities aligned with delivery objectives, improving value generation and cost efficiency.

\begin{table}[!t]
\centering
\caption{Delivery performance, API costs, and inference latency of AdCoEvo and foundation models in Sparse. Bold values indicate the best results.}
\label{tab:general-models}
\begingroup
\fontsize{7.5}{9.0}\selectfont
\setlength{\tabcolsep}{0.4pt}
\renewcommand{\arraystretch}{1.16}
\begin{tabular*}{\linewidth}{@{\extracolsep{\fill}}lccccccccc@{}}
\toprule
& \multicolumn{4}{c}{\textbf{Business Performance}} & \multicolumn{2}{c}{\textbf{Estimated API Cost}} & \multicolumn{3}{c}{\textbf{Latency}} \\
\cmidrule(lr){2-5}\cmidrule(lr){6-7}\cmidrule(l){8-10}
\multicolumn{1}{c}{{\fontsize{7.5}{9.0}\selectfont Model}} & {\fontsize{7.5}{9.0}\selectfont ADV\,$\uparrow$} & {\fontsize{7.5}{9.0}\selectfont CPA\,$\downarrow$} & {\fontsize{7.5}{9.0}\selectfont IR (\%)\,$\downarrow$} & {\fontsize{7.5}{9.0}\selectfont Score\,$\uparrow$} & {\fontsize{7.5}{9.0}\selectfont CNY/Period\,$\downarrow$} & {\fontsize{7.5}{9.0}\selectfont CNY/10K ADV\,$\downarrow$} & {\fontsize{7.5}{9.0}\selectfont P50(s)\,$\downarrow$} & {\fontsize{7.5}{9.0}\selectfont P95(s)\,$\downarrow$} & {\fontsize{7.5}{9.0}\selectfont Period(s)\,$\downarrow$} \\
\midrule
\multicolumn{10}{@{}c@{}}{\rule[-0.5ex]{0pt}{2.6ex}\textit{Closed-Source}} \\
GPT-5.6-Terra & 31,243.6 & 100.176 & \textbf{0.00} & 1.3742 & 5.733 & 1.8348 & 3.84 & 5.27 & 710.8 \\
Claude-Sonnet-5 & 31,024.5 & 100.027 & 3.22 & 1.3344 & 8.152 & 2.6275 & 9.18 & 15.30 & 1,340.0 \\
Gemini-3.1-Flash-Lite & 30,291.8 & 101.393 & 8.16 & 1.2348 & 0.683 & 0.2255 & 1.40 & 2.23 & 470.8 \\
\noalign{\vskip0.2ex
  \hbox to\linewidth{\leaders\hbox{\kern1.5pt\vrule width3pt height0.3pt\kern1.5pt}\hfill}
  \vskip0.2ex}
\multicolumn{10}{@{}c@{}}{\rule[-0.5ex]{0pt}{2.6ex}\textit{Open-Source}} \\
DeepSeek-V4-Pro & 30,890.9 & 100.147 & \textbf{0.00} & 1.3591 & 0.830 & 0.2688 & 3.96 & 6.62 & 724.4 \\
Kimi-K3 & 29,055.5 & 102.191 & 8.98 & 1.1630 & 13.441 & 4.6258 & 15.33 & 82.97 & 3,346.1 \\
GLM-5.2 & 28,056.8 & 109.419 & 8.74 & 1.0423 & 1.427 & 0.5084 & 17.86 & 34.19 & 2,170.3 \\
Qwen3.5-397B-A17B & 27,152.6 & 117.737 & 11.58 & 0.9003 & 2.334 & 0.8597 & 14.49 & 26.98 & 1,709.7 \\
\noalign{\vskip0.2ex
  \hbox to\linewidth{\leaders\hbox{\kern1.5pt\vrule width3pt height0.3pt\kern1.5pt}\hfill}
  \vskip0.2ex}
\multicolumn{10}{@{}c@{}}{\rule[-0.5ex]{0pt}{2.6ex}\textit{AdCoEvo}} \\
\textbf{AdCoEvo} & \textbf{32,408.7} & \textbf{96.936} & \textbf{0.00} & \textbf{1.4731} & \textbf{0.235} & \textbf{0.0725} & \textbf{1.20} & \textbf{2.19} & \textbf{397.1} \\
\bottomrule
\end{tabular*}
\endgroup

\end{table}

In addition to delivery performance, AdCoEvo also has lower API costs. Based on actual API usage across all roles and the corresponding listed prices,\footnote{Cost estimates use the OpenRouter prices recorded on September 1, 2026.} its average API cost is CNY 0.235 per period, 65.59\% lower than that of Gemini-3.1-Flash-Lite, the least expensive foundation model, and its cost per 10,000 ADV is CNY 0.0725. Lower API costs help scale advertising delivery.

Qwen3.5-397B-A17B, a larger model from the same Qwen family, does not achieve better delivery results, further indicating that operational capability cannot be assessed by model size alone. Organizing decisions and learning around the advertising delivery task provides another effective way to balance business performance and operational cost, beyond simply increasing the size of the backbone model.

\subsection{Ablation Analysis}
\label{sec:5-4}

\subsubsection{Role Composition Ablation}
\label{sec:5-4-1}

To examine collaboration among specialized roles, we construct combinations of Planner, Optimizer, and Inspector without domain-specific post-training. As shown in Table~\ref{tab:roles}, using Planner alone increases ADV but also raises CPA; using Optimizer alone reduces CPA but decreases ADV. Combining the two still does not improve both metrics simultaneously. Adding Inspector increases ADV by 4.95\% and reduces CPA by 9.57\% relative to the Planner--Optimizer combination. The full system outperforms No Operation on both metrics.

\begin{table}[!t]
\captionsetup{singlelinecheck=false,justification=raggedright}
\begin{minipage}[t]{0.48\textwidth}
\vspace{0pt}
\caption{ADV and CPA of different role combinations in Sparse without domain-specific post-training.}
\label{tab:roles}
\end{minipage}\hfill
\begin{minipage}[t]{0.48\textwidth}
\vspace{0pt}
\caption{Training-stage ablation in Sparse.}
\label{tab:training}
\end{minipage}
\par\nointerlineskip
\begin{minipage}[t]{0.48\textwidth}
\vspace{0pt}
\centering
\begingroup
\fontsize{7.5}{9.0}\selectfont
\setlength{\tabcolsep}{6pt}
\begin{tabular}{@{}lrr@{}}
\toprule
Method & ADV $\uparrow$ & CPA $\downarrow$ \\
\midrule
No Operation & 26,374.7 & 116.207 \\
Planner & 26,682.1 & 125.751 \\
Optimizer & 25,004.8 & 113.410 \\
P+O & 26,243.8 & 121.240 \\
P+O+I & \textbf{27,543.2} & \textbf{109.641} \\
\bottomrule
\end{tabular}
\endgroup

\end{minipage}\hfill
\begin{minipage}[t]{0.48\textwidth}
\vspace{0pt}
\centering
\begingroup
\fontsize{7.5}{9.0}\selectfont
\setlength{\tabcolsep}{6pt}
\begin{tabular}{@{}lrr@{}}
\toprule
Method & ADV $\uparrow$ & CPA $\downarrow$ \\
\midrule
No Operation & 26,374.7 & 116.207 \\
w/o Training & 27,543.2 & 109.641 \\
SFT & 30,892.4 & 100.016 \\
GSPO & 31,175.1 & 98.150 \\
AdCoEvo & \textbf{32,408.7} & \textbf{96.936} \\
\bottomrule
\end{tabular}
\endgroup

\end{minipage}
\end{table}

This change indicates that local improvements from individual roles do not simply add up to overall gains when more agents are introduced. Budget planning and ad-unit adjustment influence one another and require collaboration around a shared delivery objective. Inspector turns execution outcomes into diagnostic guidance for Planner and Optimizer, enabling both roles to revise subsequent decisions using prior delivery feedback.

\subsubsection{Policy Learning Stage Ablation}
\label{sec:5-4-2}

With all roles present, we compare four settings: before domain-specific post-training (w/o Training), SFT alone, subsequent GSPO offline reinforcement learning, and co-evolution. As shown in Table~\ref{tab:training}, SFT increases ADV and reduces CPA; subsequent GSPO further improves both metrics. Business performance improves with training even when the division of roles remains unchanged, indicating that the collaborative structure alone does not guarantee high-quality decisions; the corresponding operational capabilities also need to be developed. Supervised demonstrations provide initial decision-making capabilities, while reinforcement learning further optimizes delivery strategies using execution outcomes, making multi-agent collaboration more effective in supporting value generation and cost control.

Compared with AdCoEvo (Fixed), AdCoEvo increases ADV by 3.96\% and reduces CPA by 1.24\% in Sparse. In General, ADV increases by 4.22\% and CPA decreases by 2.21\%. This improvement shows that the operational capabilities established through offline training still have room to improve. In a changing advertising market, co-evolution enables the multi-agent system to continually improve its policies using new settled feedback, yielding greater delivery returns on top of its existing collaboration and operational capabilities.

\subsubsection{Co-Evolution Ablation}
\label{sec:5-4-3}

To study role participation and rotation, we retain all roles and compare Optimizer-only updates, Planner--Optimizer updates, and AdCoEvo against AdCoEvo (Fixed). To equalize role-update counts, the first setting updates Optimizer every period; the second alternates Planner and Optimizer across periods.

\begin{table}[!t]
\centering
\caption{ADV and Score gains (\%) over AdCoEvo (Fixed) under different role-update settings in Sparse. All settings retain the complete role composition.}
\label{tab:coevolution-ablation}
\begingroup
\fontsize{7.5}{9.0}\selectfont
\setlength{\tabcolsep}{4.5pt}
\begin{tabular}{@{}l*{6}{r}@{}}
\toprule
 & \multicolumn{2}{c}{O gains} & \multicolumn{2}{c}{P+O gains} & \multicolumn{2}{c}{AdCoEvo gains} \\
\cmidrule(lr){2-3}\cmidrule(lr){4-5}\cmidrule(l){6-7}
Seq. & ADV\,$\uparrow$ & Score\,$\uparrow$ & ADV\,$\uparrow$ & Score\,$\uparrow$ & ADV\,$\uparrow$ & Score\,$\uparrow$ \\
\midrule
1 & 0.62 & 2.25 & 0.93 & 2.56 & \textbf{1.28} & \textbf{2.91} \\
2 & 0.74 & 3.30 & 1.03 & 3.58 & \textbf{5.34} & \textbf{7.93} \\
3 & 0.64 & 2.84 & 1.01 & 3.11 & \textbf{6.89} & \textbf{9.10} \\
4 & 0.56 & 1.67 & 0.94 & 1.98 & \textbf{2.10} & \textbf{3.02} \\
5 & 0.59 & 2.11 & 1.33 & 2.17 & \textbf{4.32} & \textbf{5.66} \\
\midrule
Overall & 0.63 & 2.08 & 1.05 & 2.41 & \textbf{3.96} & \textbf{5.26} \\
\bottomrule
\end{tabular}
\endgroup

\end{table}

As shown in Table~\ref{tab:coevolution-ablation}, full co-evolution achieves higher ADV and composite scores than updating Planner and Optimizer across all five market sequences. Overall, ADV further increases by 2.88\% and the composite score by 2.78\%, while CPA increases by only approximately 0.10\%. This shows that, while retaining diagnostic guidance, allowing Inspector to participate in co-evolution can further increase advertiser value at a similar cost per acquisition, yielding higher delivery returns.

\begin{figure}[!t]
\centering
\includegraphics[width=\linewidth]{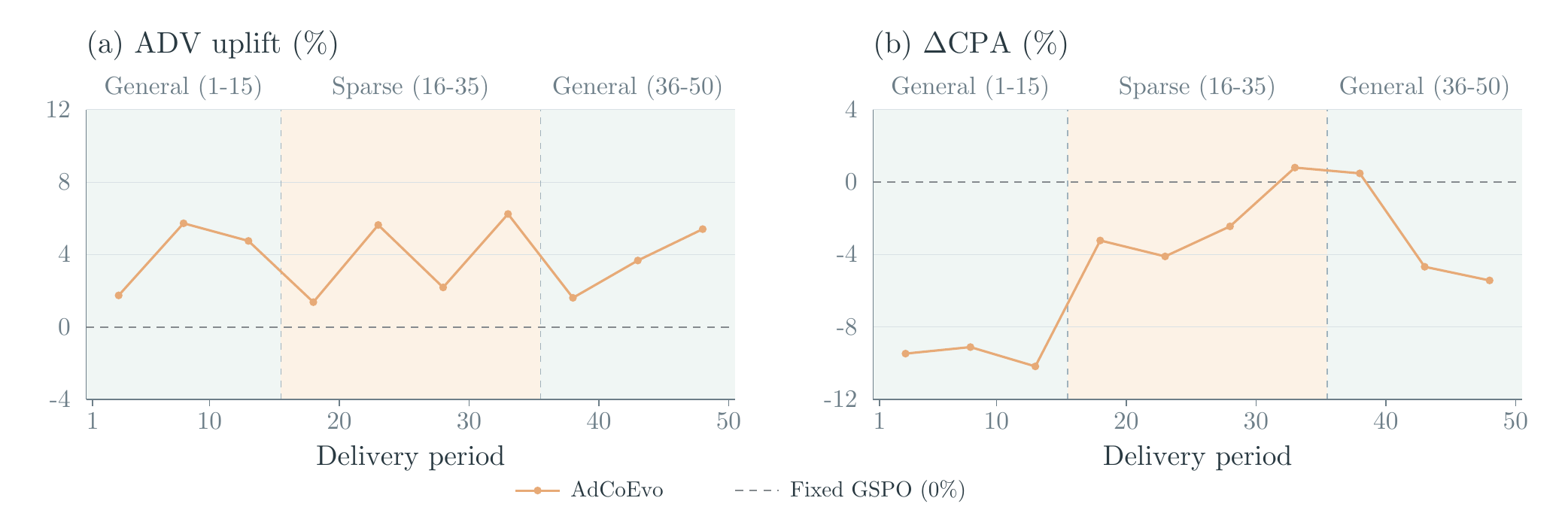}
\caption{ADV uplift and CPA change relative to AdCoEvo (Fixed) under market switches, averaged across five market sequences using five-period windows.}
\label{fig:market-switch}
\end{figure}

To examine how co-evolution adapts to market changes, we switch from General to Sparse at period 16 and back to General at period 36, comparing AdCoEvo with AdCoEvo (Fixed). We report results averaged over five market sequences using five-period windows. As shown in Figure~\ref{fig:market-switch}, AdCoEvo retains an ADV advantage after the market switches and gradually recovers its CPA improvement after the return to General, showing that co-evolution can improve adaptation to market changes.

\subsubsection{Diagnostic Guidance Ablation}
\label{sec:5-4-4}

To examine the role of diagnostic guidance, we replace Inspector-generated diagnoses with fixed statistical summaries and compare delivery performance under the two feedback settings. Relative to fixed statistical summaries, Inspector-generated diagnoses increase ADV by 1.61\% and reduce CPA by 1.55\% in Sparse, with corresponding improvements of 1.64\% and 1.34\% in General. This shows that converting delivery outcomes into diagnostic guidance better supports subsequent account planning and strategy adjustment than providing statistical information alone.

\subsection{Budget Constraints and Parameter Sensitivity}
\label{sec:5-5}

\begin{figure}[!t]
\centering
\includegraphics[width=\linewidth]{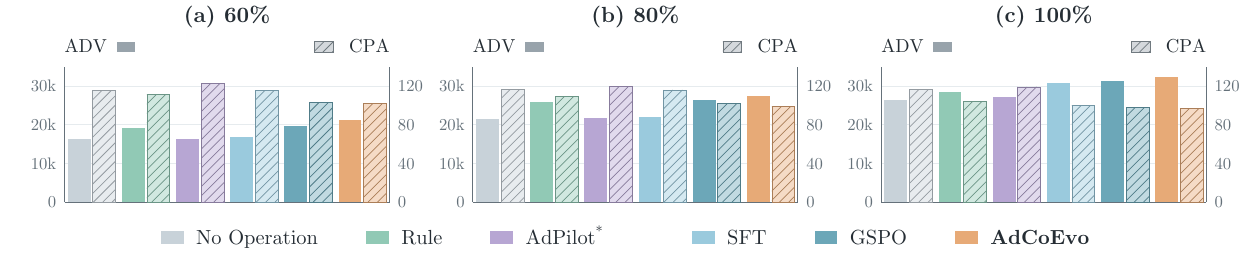}
\caption{ADV (solid bars, left axis) and CPA (hatched bars, right axis) under different initial budgets in Sparse.}
\label{fig:budget}
\end{figure}

To examine whether AdCoEvo's delivery gains depend on a particular resource setting, we compare business performance under different initial budgets. As shown in Figure~\ref{fig:budget}, AdCoEvo achieves the highest ADV and lowest CPA at 60\%, 80\%, and 100\% of the original budget. Even when the budget is reduced to 60\%, it increases ADV by 7.17\% and reduces CPA by 0.78\% relative to AdCoEvo (Fixed). Tighter budgets require the system to coordinate account planning and ad-unit adjustment with more limited resources, yet AdCoEvo still improves both value generation and cost efficiency. Thus, co-evolution retains its gains over the fixed offline policy at lower budgets.

In addition to budget, we examine the observation length $W$ and the maximum number of tCPA adjustments per ad unit per period $K$. As shown in Table~\ref{tab:parameters}, the default setting of $W=12$ and $K=8$ achieves the highest ADV and lowest CPA among the configurations compared. Extending $W$ from 6 to 12 improves both metrics, whereas extending it further to 18 reduces performance. A longer observation period provides more information before decisions are made but shortens the time during which planning and adjustment can affect subsequent delivery, requiring a balance between observation and execution.

\begin{table}[!t]
\centering
\caption{Effects of observation length $W$ and adjustment limit $K$ on delivery performance.}
\label{tab:parameters}
\begingroup
\fontsize{7.5}{9.0}\selectfont
\setlength{\tabcolsep}{2pt}
\begin{tabular*}{0.60\linewidth}{@{\extracolsep{\fill}}lccc@{}}
\toprule
\multicolumn{1}{c}{\textbf{Parameter}} & \multicolumn{1}{c}{\textbf{ADV $\uparrow$}} & \multicolumn{1}{c}{\textbf{CPA $\downarrow$}} & \multicolumn{1}{c}{\textbf{Score $\uparrow$}} \\
\midrule
\multicolumn{4}{@{}c@{}}{\rule[-0.5ex]{0pt}{2.6ex}\textit{Observation length}} \\
$W = 6$ & 29,547.0 & 97.742 & 1.3319 \\
$W = 12$ (default) & \textbf{32,408.7} & \textbf{96.936} & \textbf{1.4731} \\
$W = 18$ & 30,649.6 & 107.914 & 1.2514 \\
\noalign{\vskip0.2ex
  \hbox to0.60\linewidth{\leaders\hbox{\kern1.5pt\vrule width3pt height0.3pt\kern1.5pt}\hfill}
  \vskip0.2ex}
\multicolumn{4}{@{}c@{}}{\rule[-0.5ex]{0pt}{2.6ex}\textit{Adjustment limit}} \\
$K = 4$ & 31,992.6 & 97.240 & 1.4496 \\
$K = 8$ (default) & \textbf{32,408.7} & \textbf{96.936} & \textbf{1.4731} \\
$K = 12$ & 31,268.6 & 99.510 & 1.3845 \\
\bottomrule
\end{tabular*}
\endgroup

\end{table}

Reducing $K$ from 8 to 4 decreases ADV by 1.28\% and raises CPA by 0.31\%, while performance remains close to the default configuration under the tighter adjustment constraint. Relaxing the limit to 12 lowers ADV and raises CPA, so more adjustment opportunities do not necessarily improve performance.

\FloatBarrier

\section{Discussion and Conclusion}
\label{sec:6}

We present AdCoEvo, a multi-agent system for advertising delivery that organizes delivery decisions through role specialization across different timescales and operational granularities, and improves delivery policies through offline development of operational capabilities and co-evolution. Experiments show that operational-level decisions, role collaboration, and post-training all help improve delivery performance, while co-evolution provides further gains beyond offline reinforcement learning.

However, gains vary across environments, and the effectiveness of co-evolution still needs evaluation in specific delivery scenarios. How operational roles' decision frequencies and scopes of influence affect co-evolution, and how advantage estimation and update mechanisms can be improved accordingly, warrant further study to help multi-agent systems use actual delivery feedback more effectively.

\begingroup
\addtolength{\bibsep}{-1pt}
\bibliographystyle{unsrtnat}
\bibliography{references}

\begin{thebibliography}{43}
\providecommand{\natexlab}[1]{#1}
\providecommand{\url}[1]{\texttt{#1}}
\expandafter\ifx\csname urlstyle\endcsname\relax
  \providecommand{\doi}[1]{doi: #1}\else
  \providecommand{\doi}{doi: \begingroup \urlstyle{rm}\Url}\fi

\bibitem[Zhang et~al.(2014)Zhang, Yuan, and Wang]{zhang2014ortb}
Weinan Zhang, Shuai Yuan, and Jun Wang.
\newblock Optimal real-time bidding for display advertising.
\newblock In \emph{Proceedings of the 20th ACM SIGKDD International Conference on Knowledge Discovery and Data Mining}, pages 1077--1086. Association for Computing Machinery, 2014.
\newblock \doi{10.1145/2623330.2623633}.
\newblock URL \url{https://doi.org/10.1145/2623330.2623633}.

\bibitem[Sun et~al.(2026)Sun, He, Li, et~al.]{sun2026adpilot}
Shuoshuo Sun, Qi~He, Xiaoting Li, et~al.
\newblock {AdPilot}: Towards fully autonomous advertising delivery via agentic reinforcement learning.
\newblock In \emph{Proceedings of the 32nd ACM SIGKDD Conference on Knowledge Discovery and Data Mining V.2}, pages 8021--8031, 2026.
\newblock \doi{10.1145/3770855.3818315}.
\newblock URL \url{https://doi.org/10.1145/3770855.3818315}.

\bibitem[Mou et~al.(2022)]{mou2022sorl}
Zhiyu Mou et~al.
\newblock Sustainable online reinforcement learning for auto-bidding.
\newblock In \emph{Advances in Neural Information Processing Systems}, volume~35, pages 2651--2663. Curran Associates, Inc., 2022.
\newblock \doi{10.52202/068431-0192}.
\newblock URL \url{https://proceedings.neurips.cc/paper_files/paper/2022/file/11faf17bf7e5412d9cded369f97db23d-Paper-Conference.pdf}.

\bibitem[Wang et~al.(2024)]{wang2024hibid}
Hao Wang et~al.
\newblock {HiBid}: A cross-channel constrained bidding system with budget allocation by hierarchical offline deep reinforcement learning.
\newblock \emph{IEEE Transactions on Computers}, 73\penalty0 (3):\penalty0 815--828, 2024.
\newblock \doi{10.1109/tc.2023.3343111}.
\newblock URL \url{https://doi.org/10.1109/tc.2023.3343111}.

\bibitem[Nuara et~al.(2018)Nuara, Trov{\`o}, Gatti, and Restelli]{nuara2018adcomb}
Alessandro Nuara, Francesco Trov{\`o}, Nicola Gatti, and Marcello Restelli.
\newblock A combinatorial-bandit algorithm for the online joint bid/budget optimization of pay-per-click advertising campaigns.
\newblock \emph{Proceedings of the AAAI Conference on Artificial Intelligence}, 32\penalty0 (1):\penalty0 2379--2386, 2018.
\newblock \doi{10.1609/aaai.v32i1.11888}.
\newblock URL \url{https://ojs.aaai.org/index.php/AAAI/article/view/11888}.

\bibitem[Song et~al.(2026)Song, Huo, Zhou, Yin, Xiao, Long, Zhang, and Yu]{song2026dara}
Mingxuan Song, Yusen Huo, Bohan Zhou, Shenglin Yin, Zhen Xiao, Jieyi Long, Zhilin Zhang, and Chuan Yu.
\newblock {DARA}: Few-shot budget allocation in online advertising via in-context decision making with {RL}-finetuned {LLM}s.
\newblock In \emph{Proceedings of the ACM Web Conference 2026}, pages 40--50. Association for Computing Machinery, 2026.
\newblock \doi{10.1145/3774904.3792129}.
\newblock URL \url{https://doi.org/10.1145/3774904.3792129}.

\bibitem[Croissant et~al.(2020)Croissant, Abeille, and Calauzenes]{croissant2020realtime}
Lorenzo Croissant, Marc Abeille, and Clement Calauzenes.
\newblock Real-time optimisation for online learning in auctions.
\newblock In \emph{Proceedings of the 37th International Conference on Machine Learning}, volume 119 of \emph{Proceedings of Machine Learning Research}, pages 2217--2226. PMLR, 2020.
\newblock URL \url{https://proceedings.mlr.press/v119/croissant20a.html}.

\bibitem[Badanidiyuru(2026)]{badanidiyuru2026monotonicity}
Ashwinkumar Badanidiyuru.
\newblock Model monotonicity in autobidding auctions: When do better predictions lead to better outcomes?
\newblock In \emph{Proceedings of the 43rd International Conference on Machine Learning}, volume 306 of \emph{Proceedings of Machine Learning Research}, pages 4802--4819. PMLR, 2026.
\newblock URL \url{https://proceedings.mlr.press/v306/badanidiyuru26a.html}.

\bibitem[{Google Ads}(n.d)]{googleads_average_cpa}
{Google Ads}.
\newblock Average {CPA}: Definition.
\newblock Google Ads Help, n.d.
\newblock URL \url{https://support.google.com/google-ads/answer/6396841?hl=en}.
\newblock Accessed: 2026-10-11.

\bibitem[Wu et~al.(2018)Wu, Chen, Yang, Wang, Tan, Zhang, Xu, and Gai]{wu2018drlb}
Di~Wu, Xiujun Chen, Xun Yang, Hao Wang, Qing Tan, Xiaoxun Zhang, Jian Xu, and Kun Gai.
\newblock Budget constrained bidding by model-free reinforcement learning in display advertising.
\newblock In \emph{Proceedings of the 27th ACM International Conference on Information and Knowledge Management}, pages 1443--1451. Association for Computing Machinery, 2018.
\newblock \doi{10.1145/3269206.3271748}.
\newblock URL \url{https://doi.org/10.1145/3269206.3271748}.

\bibitem[Cai et~al.(2017)Cai, Ren, Zhang, Malialis, Wang, Yu, and Guo]{cai2017rlb}
Han Cai, Kan Ren, Weinan Zhang, Kleanthis Malialis, Jun Wang, Yong Yu, and Defeng Guo.
\newblock Real-time bidding by reinforcement learning in display advertising.
\newblock In \emph{Proceedings of the Tenth ACM International Conference on Web Search and Data Mining}, pages 661--670. Association for Computing Machinery, 2017.
\newblock \doi{10.1145/3018661.3018702}.
\newblock URL \url{https://doi.org/10.1145/3018661.3018702}.

\bibitem[Guo et~al.(2024)]{guo2024aigb}
Jiayan Guo et~al.
\newblock Generative auto-bidding via conditional diffusion modeling.
\newblock In \emph{Proceedings of the 30th ACM SIGKDD Conference on Knowledge Discovery and Data Mining}, pages 5038--5049, 2024.
\newblock \doi{10.1145/3637528.3671526}.
\newblock URL \url{https://doi.org/10.1145/3637528.3671526}.

\bibitem[Cai et~al.(2025)]{cai2025rtbagent}
Leng Cai et~al.
\newblock {RTBAgent}: A {LLM}-based agent system for real-time bidding.
\newblock In \emph{Companion Proceedings of the ACM on Web Conference 2025}, pages 104--113, 2025.
\newblock \doi{10.1145/3701716.3715259}.
\newblock URL \url{https://doi.org/10.1145/3701716.3715259}.

\bibitem[Su et~al.(2024)Su, Huo, Zhang, et~al.]{auctionnet2024}
Kefan Su, Yusen Huo, Zhilin Zhang, et~al.
\newblock {AuctionNet}: A novel benchmark for decision-making in large-scale games.
\newblock In \emph{Advances in Neural Information Processing Systems}, volume~37, pages 94428--94452. Curran Associates, Inc., 2024.
\newblock \doi{10.52202/079017-2993}.
\newblock URL \url{https://proceedings.neurips.cc/paper_files/paper/2024/file/ab9b7c23edfea0011507f7e1eae82cd2-Paper-Datasets_and_Benchmarks_Track.pdf}.

\bibitem[Li et~al.(2023)Li, Hammoud, Itani, et~al.]{li2023camel}
Guohao Li, Hasan Abed Al~Kader Hammoud, Hani Itani, et~al.
\newblock {CAMEL}: Communicative agents for ``mind'' exploration of large language model society.
\newblock In \emph{Advances in Neural Information Processing Systems}, volume~36, pages 51991--52008, 2023.
\newblock \doi{10.52202/075280-2264}.
\newblock URL \url{https://proceedings.neurips.cc/paper_files/paper/2023/hash/a3621ee907def47c1b952ade25c67698-Abstract-Conference.html}.

\bibitem[Wu et~al.(2024)Wu, Bansal, Zhang, et~al.]{wu2024autogen}
Qingyun Wu, Gagan Bansal, Jieyu Zhang, et~al.
\newblock {AutoGen}: Enabling next-gen {LLM} applications via multi-agent conversations.
\newblock In \emph{First Conference on Language Modeling}, 2024.
\newblock URL \url{https://openreview.net/forum?id=BAakY1hNKS}.

\bibitem[Hong et~al.(2024)Hong, Zhuge, Chen, et~al.]{hong2024metagpt}
Sirui Hong, Mingchen Zhuge, Jonathan Chen, et~al.
\newblock {MetaGPT}: Meta programming for a multi-agent collaborative framework.
\newblock In \emph{The Twelfth International Conference on Learning Representations}, pages 23247--23275, 2024.
\newblock URL \url{https://proceedings.iclr.cc/paper_files/paper/2024/hash/6507b115562bb0a305f1958ccc87355a-Abstract-Conference.html}.

\bibitem[Qian et~al.(2024)Qian, Liu, Liu, et~al.]{qian2024chatdev}
Chen Qian, Wei Liu, Hongzhang Liu, et~al.
\newblock {ChatDev}: Communicative agents for software development.
\newblock In \emph{Proceedings of the 62nd Annual Meeting of the Association for Computational Linguistics (Volume 1: Long Papers)}, pages 15174--15186, Bangkok, Thailand, August 2024. Association for Computational Linguistics.
\newblock \doi{10.18653/v1/2024.acl-long.810}.
\newblock URL \url{https://aclanthology.org/2024.acl-long.810/}.

\bibitem[Tao et~al.(2024)Tao, Zhou, Wang, Zhang, Zhang, and Cheng]{tao2024magis}
Wei Tao, Yucheng Zhou, Yanlin Wang, Wenqiang Zhang, Hongyu Zhang, and Yu~Cheng.
\newblock {MAGIS}: {LLM}-based multi-agent framework for {GitHub} issue resolution.
\newblock In \emph{Advances in Neural Information Processing Systems}, volume~37, pages 51963--51993, 2024.
\newblock \doi{10.52202/079017-1647}.
\newblock URL \url{https://proceedings.neurips.cc/paper_files/paper/2024/hash/5d1f02132ef51602adf07000ca5b6138-Abstract-Conference.html}.

\bibitem[Niu et~al.(2025)Niu, Song, Lian, Shen, Yao, Zhang, and Liu]{niu2025flow}
Boye Niu, Yiliao Song, Kai Lian, Yifan Shen, Yu~Yao, Kun Zhang, and Tongliang Liu.
\newblock {Flow: Modularized Agentic Workflow Automation}.
\newblock In \emph{International Conference on Learning Representations}, pages 74949--74977, 2025.
\newblock URL \url{https://proceedings.iclr.cc/paper_files/paper/2025/hash/ba84da6921f3040b74ee163aa7451f53-Abstract-Conference.html}.

\bibitem[Zhou et~al.(2026)Zhou, Wan, Sun, Palangi, Iqbal, Vuli{\'c}, Korhonen, and Arik]{zhou2026mass}
Han Zhou, Xingchen Wan, Ruoxi Sun, Hamid Palangi, Shariq Iqbal, Ivan Vuli{\'c}, Anna Korhonen, and Sercan Arik.
\newblock {Multi-Agent Design: Optimizing Agents with Better Prompts and Topologies}.
\newblock In \emph{International Conference on Learning Representations}, pages 15844--15872, 2026.
\newblock URL \url{https://proceedings.iclr.cc/paper_files/paper/2026/hash/1ab4e0e8f35078109bc78d7b465d306f-Abstract-Conference.html}.

\bibitem[Lin et~al.(2026)Lin, Chen, Fang, Wang, and Lin]{lin2026supervisoragent}
Fulin Lin, Shaowen Chen, Ruishan Fang, Hongwei Wang, and Tao Lin.
\newblock {Stop Wasting Your Tokens: Towards Efficient Runtime Multi-Agent Systems}.
\newblock In \emph{International Conference on Learning Representations}, pages 113497--113528, 2026.
\newblock URL \url{https://proceedings.iclr.cc/paper_files/paper/2026/hash/b86a195e70f27017c514fa0e5f80595f-Abstract-Conference.html}.

\bibitem[Chen et~al.(2024)Chen, Su, Zuo, Yang, Yuan, Chan, Yu, Lu, Hung, Qian, Qin, Cong, Xie, Liu, Sun, and Zhou]{chen2024agentverse}
Weize Chen, Yusheng Su, Jingwei Zuo, Cheng Yang, Chenfei Yuan, Chi-Min Chan, Heyang Yu, Yaxi Lu, Yi-Hsin Hung, Chen Qian, Yujia Qin, Xin Cong, Ruobing Xie, Zhiyuan Liu, Maosong Sun, and Jie Zhou.
\newblock {AgentVerse}: Facilitating multi-agent collaboration and exploring emergent behaviors.
\newblock In \emph{International Conference on Learning Representations}, pages 20094--20136, 2024.
\newblock URL \url{https://proceedings.iclr.cc/paper_files/paper/2024/hash/578e65cdee35d00c708d4c64bce32971-Abstract-Conference.html}.

\bibitem[Motwani et~al.(2025)Motwani, Smith, Das, et~al.]{motwani2025malt}
Sumeet~Ramesh Motwani, Chandler Smith, Rocktim~Jyoti Das, et~al.
\newblock {MALT}: Improving reasoning with multi-agent {LLM} training.
\newblock In \emph{Second Conference on Language Modeling}, 2025.
\newblock URL \url{https://openreview.net/forum?id=jXP9bgFack}.

\bibitem[Xue et~al.(2026)Xue, Zhou, Zhang, Zhang, Li, Zhang, Yin, Torr, Ouyang, and Bai]{xue2026comas}
Xiangyuan Xue, Yifan Zhou, Guibin Zhang, Zaibin Zhang, Yijiang Li, Chen Zhang, Zhenfei Yin, Philip Torr, Wanli Ouyang, and Lei Bai.
\newblock {CoMAS: Co-Evolving Multi-Agent Systems via Interaction Rewards}.
\newblock In \emph{International Conference on Learning Representations}, pages 32373--32394, 2026.
\newblock URL \url{https://proceedings.iclr.cc/paper_files/paper/2026/hash/372593bd318ad8b34b3a8da77e20272b-Abstract-Conference.html}.

\bibitem[Jin et~al.(2025)Jin, Zeng, Yue, Yoon, Ar{\i}k, Wang, Zamani, and Han]{jin2025searchr1}
Bowen Jin, Hansi Zeng, Zhenrui Yue, Jinsung Yoon, Sercan~{\"O}. Ar{\i}k, Dong Wang, Hamed Zamani, and Jiawei Han.
\newblock {Search-R1}: Training {LLMs} to reason and leverage search engines with reinforcement learning.
\newblock In \emph{Second Conference on Language Modeling}, 2025.
\newblock URL \url{https://openreview.net/forum?id=Rwhi91ideu}.

\bibitem[Du et~al.(2026)Du, Gong, Ling, Liu, Shen, Yao, Xu, Shi, Yang, and Chen]{du2026codegym}
Weihua Du, Hailei Gong, Zhan Ling, Kang Liu, Lingfeng Shen, Xuesong Yao, Yufei Xu, Dingyuan Shi, Yiming Yang, and Jiecao Chen.
\newblock {Generalizable End-to-End Tool-Use RL with Synthetic CodeGym}.
\newblock In \emph{International Conference on Learning Representations}, pages 17733--17756, 2026.
\newblock URL \url{https://proceedings.iclr.cc/paper_files/paper/2026/hash/1e4322fddd833f83c855660ac65e428d-Abstract-Conference.html}.

\bibitem[Qi et~al.(2025)Qi, Liu, Iong, Lai, Sun, Sun, Yang, Yang, Yao, Xu, Tang, and Dong]{qi2025webrl}
Zehan Qi, Xiao Liu, Iat~Long Iong, Hanyu Lai, Xueqiao Sun, Jiadai Sun, Xinyue Yang, Yu~Yang, Shuntian Yao, Wei Xu, Jie Tang, and Yuxiao Dong.
\newblock {WebRL}: Training {LLM} web agents via self-evolving online curriculum reinforcement learning.
\newblock In \emph{International Conference on Learning Representations}, pages 79791--79821, 2025.
\newblock URL \url{https://proceedings.iclr.cc/paper_files/paper/2025/hash/c66e1fcc9691aae706250638f36f681b-Abstract-Conference.html}.

\bibitem[Feng et~al.(2026)Feng, Huang, Qu, Zhang, Qin, Zhong, Jiang, Chi, and Zhong]{feng2026retool}
Jiazhan Feng, Shijue Huang, Xingwei Qu, Ge~Zhang, Yujia Qin, Baoquan Zhong, Chengquan Jiang, Jinxin Chi, and Wanjun Zhong.
\newblock {ReTool: Reinforcement Learning for Strategic Tool Use in LLMs}.
\newblock In \emph{International Conference on Learning Representations}, pages 37909--37926, 2026.
\newblock URL \url{https://proceedings.iclr.cc/paper_files/paper/2026/hash/4038c9208dfc22644c60ad39c24e5c53-Abstract-Conference.html}.

\bibitem[Li et~al.(2026)Li, Zhang, Han, Liu, Xie, Zhang, Choi, Zou, and Lu]{li2026agentflow}
Zhuofeng Li, Haoxiang Zhang, Seungju Han, Sheng Liu, Jianwen Xie, Yu~Zhang, Yejin Choi, James~Y Zou, and Pan Lu.
\newblock {In-the-Flow Agentic System Optimization for Effective Planning and Tool Use}.
\newblock In \emph{International Conference on Learning Representations}, pages 50524--50570, 2026.
\newblock URL \url{https://proceedings.iclr.cc/paper_files/paper/2026/hash/52c4a736cd63b3c7bf52011a1f66f3f6-Abstract-Conference.html}.

\bibitem[Zhang et~al.(2026)Zhang, Tian, Liu, Zeng, Zhu, Jia, Fan, Lv, Zuo, Jiang, Wang, Wang, Hua, Long, Gao, Sun, Ma, Cui, Ding, Qi, and Zhou]{zhang2026marti}
Kaiyan Zhang, Kai Tian, Runze Liu, Sihang Zeng, Xuekai Zhu, Guoli Jia, Yuchen Fan, Xingtai Lv, Yuxin Zuo, Che Jiang, Yuru Wang, Jianyu Wang, Ermo Hua, Xinwei Long, Junqi Gao, Youbang Sun, Zhiyuan Ma, Ganqu Cui, Ning Ding, Biqing Qi, and Bowen Zhou.
\newblock {MARTI: A Framework for Multi-Agent LLM Systems Reinforced Training and Inference}.
\newblock In \emph{International Conference on Learning Representations}, pages 138790--138813, 2026.
\newblock URL \url{https://proceedings.iclr.cc/paper_files/paper/2026/hash/e045c55a71fb4fc04d9a20579415247d-Abstract-Conference.html}.

\bibitem[Ouyang et~al.(2022)Ouyang, Wu, Jiang, Almeida, Wainwright, Mishkin, Zhang, Agarwal, Slama, Ray, Schulman, Hilton, Kelton, Miller, Simens, Askell, Welinder, Christiano, Leike, and Lowe]{ouyang2022instructgpt}
Long Ouyang, Jeffrey Wu, Xu~Jiang, Diogo Almeida, Carroll Wainwright, Pamela Mishkin, Chong Zhang, Sandhini Agarwal, Katarina Slama, Alex Ray, John Schulman, Jacob Hilton, Fraser Kelton, Luke Miller, Maddie Simens, Amanda Askell, Peter Welinder, Paul~F. Christiano, Jan Leike, and Ryan Lowe.
\newblock Training language models to follow instructions with human feedback.
\newblock In \emph{Advances in Neural Information Processing Systems}, volume~35, pages 27730--27744. Curran Associates, Inc., 2022.
\newblock \doi{10.52202/068431-2011}.
\newblock URL \url{https://proceedings.neurips.cc/paper_files/paper/2022/hash/b1efde53be364a73914f58805a001731-Abstract-Conference.html}.

\bibitem[Zeng et~al.(2024)Zeng, Liu, Lu, Wang, Liu, Dong, and Tang]{zeng2024agenttuning}
Aohan Zeng, Mingdao Liu, Rui Lu, Bowen Wang, Xiao Liu, Yuxiao Dong, and Jie Tang.
\newblock {AgentTuning}: Enabling generalized agent abilities for {LLMs}.
\newblock In \emph{Findings of the Association for Computational Linguistics: ACL 2024}, pages 3053--3077. Association for Computational Linguistics, 2024.
\newblock \doi{10.18653/v1/2024.findings-acl.181}.
\newblock URL \url{https://aclanthology.org/2024.findings-acl.181/}.

\bibitem[Ahmadian et~al.(2024)Ahmadian, Cremer, Gall{\'e}, Fadaee, Kreutzer, Pietquin, {\"U}st{\"u}n, and Hooker]{ahmadian2024rloo}
Arash Ahmadian, Chris Cremer, Matthias Gall{\'e}, Marzieh Fadaee, Julia Kreutzer, Olivier Pietquin, Ahmet {\"U}st{\"u}n, and Sara Hooker.
\newblock Back to basics: Revisiting {REINFORCE}-style optimization for learning from human feedback in {LLM}s.
\newblock In \emph{Proceedings of the 62nd Annual Meeting of the Association for Computational Linguistics (Volume 1: Long Papers)}, pages 12248--12267. Association for Computational Linguistics, 2024.
\newblock \doi{10.18653/v1/2024.acl-long.662}.
\newblock URL \url{https://aclanthology.org/2024.acl-long.662/}.

\bibitem[Zheng et~al.(2025)]{zheng2025gspo}
Chujie Zheng et~al.
\newblock Group sequence policy optimization.
\newblock arXiv preprint, 2025.
\newblock URL \url{https://arxiv.org/abs/2507.18071}.

\bibitem[Hou et~al.(2026)]{hou2026sao}
Zhenyu Hou et~al.
\newblock Single-rollout asynchronous optimization for agentic reinforcement learning.
\newblock arXiv preprint, 2026.
\newblock URL \url{https://arxiv.org/abs/2607.07508}.

\bibitem[Schulman et~al.(2016)Schulman, Moritz, Levine, Jordan, and Abbeel]{schulman2016gae}
John Schulman, Philipp Moritz, Sergey Levine, Michael~I. Jordan, and Pieter Abbeel.
\newblock High-dimensional continuous control using generalized advantage estimation.
\newblock In \emph{International Conference on Learning Representations}, 2016.
\newblock URL \url{https://arxiv.org/abs/1506.02438}.

\bibitem[{DeepSeek-AI}(2026)]{deepseekai2026deepseekv4}
{DeepSeek-AI}.
\newblock {DeepSeek-V4}: Towards highly efficient million-token context intelligence.
\newblock Technical report, 2026.
\newblock URL \url{https://arxiv.org/abs/2606.19348}.

\bibitem[{Kimi Team}(2026)]{kimiteam2026kimik3}
{Kimi Team}.
\newblock {Kimi K3}: Open frontier intelligence.
\newblock arXiv preprint, 2026.
\newblock URL \url{https://arxiv.org/abs/2607.24653}.

\bibitem[{Z.AI}(2026)]{zai2026glm52}
{Z.AI}.
\newblock {GLM-5.2}: Built for long-horizon tasks, June 2026.
\newblock URL \url{https://huggingface.co/blog/zai-org/glm-52-blog}.

\bibitem[{Qwen Team}(2026)]{qwen2026qwen35}
{Qwen Team}.
\newblock {Qwen3.5}: Towards native multimodal agents, February 2026.
\newblock URL \url{https://qwen.ai/blog?id=qwen3.5}.

\bibitem[Yang et~al.(2025)Yang, Li, Yang, et~al.]{yang2025qwen3}
An~Yang, Anfeng Li, Baosong Yang, et~al.
\newblock {Qwen3} technical report.
\newblock arXiv preprint, 2025.
\newblock URL \url{https://arxiv.org/abs/2505.09388}.

\bibitem[Hu et~al.(2022)Hu, Shen, Wallis, Allen-Zhu, Li, Wang, Wang, and Chen]{hu2022lora}
Edward~J. Hu, Yelong Shen, Phillip Wallis, Zeyuan Allen-Zhu, Yuanzhi Li, Shean Wang, Lu~Wang, and Weizhu Chen.
\newblock {LoRA}: Low-rank adaptation of large language models.
\newblock In \emph{International Conference on Learning Representations}, 2022.
\newblock URL \url{https://openreview.net/forum?id=nZeVKeeFYf9}.

\end{thebibliography}
\endgroup

\clearpage
\appendix
\section{Role Reward Computation}
\label{sec:reward-details}

\setcounter{equation}{0}
\renewcommand{\theequation}{A.\arabic{equation}}
\renewcommand{\theHequation}{A.\arabic{equation}}

We detail aggregation scopes and parameters for offline and co-evolution rewards (Section~\ref{sec:4-2}).

\subsection{Rewards for Planner and Optimizer}
\label{sec:reward-po}

Equations~\eqref{eq:coevo-reward-score}--\eqref{eq:role-rewards} combine separate returns for the current account and all controlled accounts. Planner uses results from observation end to period end; Optimizer uses current-step increments. Each term aggregates ADV, spend, and conversions $F,C,N$ in its scope to compute the scaled proxy return.

The CPA target averages initial tCPA across all units, weighted by conversions. For settlement window $\mathcal W$, let $d_u$ be unit $u$'s initial tCPA and $N_{u,\mathcal W}$ its conversions. The target cost allowance $M_{\mathcal W}=\sum_ud_uN_{u,\mathcal W}$ and total conversions $N_{\mathcal W}=\sum_uN_{u,\mathcal W}$ give $\rho=M_{\mathcal W}/N_{\mathcal W}$; when $N_{\mathcal W}=0$, we set $\rho=0$.

Both Sparse and General use this computation. Training sets $\eta=1$, $s=10000$, and $\omega=0.2$ to control the cost-deviation weight, reward scaling, and overall reward weight, respectively. Planner's reward settles at period end; Optimizer's current-step reward is used in the advantage recursion in Appendix~\ref{sec:gae-details}.

\subsection{Period-Level Feedback for Inspector}
\label{sec:reward-inspector}

Period feedback aggregates value generation, conversions, cost control, and adjustment execution across all controlled accounts. Let $F_e,C_e,N_e$ denote aggregate ADV, spend, and conversions in period $e$. The feedback vector and fixed weights in Eq.~\eqref{eq:role-rewards} are

\begin{equation}
\begin{aligned}
\mathbf y_e&=\bigl[\widetilde F_e,\widetilde N_e,m_e^{\mathrm{CPA}},
-r_e^{\mathrm{forced}}\bigr]^\top,\\
\mathbf w&=[1.0,0.2,0.5,0.1]^\top.
\end{aligned}
\label{eq:inspector-feedback}
\end{equation}

To obtain $\widetilde F_e$ and $\widetilde N_e$, we subtract training-set means $\mu_F,\mu_N$ from ADV and conversions, divide by the corresponding standard deviations $\sigma_F,\sigma_N$ plus a numerical stabilizer $\epsilon>0$, and clip to $[-3,3]$. The means and standard deviations remain frozen during training.

The CPA constraint margin compares actual spending with the target cost allowance $M_e=\sum_ud_uN_{e,u}$, where $u$ ranges over all controlled units and $N_{e,u}$ counts unit $u$'s conversions in period $e$:

\begin{equation}
m_e^{\mathrm{CPA}}=\operatorname{clip}\!\left(
\frac{M_e-C_e}{\max(M_e,\epsilon)},-3,3\right).
\label{eq:cpa-margin}
\end{equation}

This component is positive below the target cost allowance and negative above it.

The forced-HOLD rate reflects adjustment requests made after the adjustment limit has been reached. Let $H_e$ count SET requests converted to HOLD because the limit has been reached, and $S_e$ count all SET requests:

\begin{equation}
r_e^{\mathrm{forced}}=\operatorname{clip}\!\left(
\frac{H_e}{\max(S_e,1)},0,3\right).
\label{eq:action-feedback}
\end{equation}

This component enters the feedback vector negatively. Once period $e$'s feedback is settled, the system computes the previous diagnosis's reward as the weighted change between adjacent periods, $r_{e-1}^I=\mathbf w^\top(\mathbf y_e-\mathbf y_{e-1})$.

\FloatBarrier
\section{Offline Optimization Details and Algorithm}
\label{sec:offline-details}

\setcounter{equation}{0}
\renewcommand{\theequation}{B.\arabic{equation}}
\renewcommand{\theHequation}{B.\arabic{equation}}
\setcounter{algocf}{0}
\renewcommand{\thealgocf}{B.\arabic{algocf}}
\providecommand{\theHalgocf}{}
\renewcommand{\theHalgocf}{B.\arabic{algocf}}

\subsection{Candidate Advantages and Sequence-Level Optimization}
\label{sec:offline-optimization}

At the same decision snapshot, the system fixes the policies of the other roles, samples candidate responses for the target role using behavior policy $\pi_b^r$, and computes returns according to the reward definitions in Appendix~\ref{sec:reward-details}. Optimizer's candidate return is the undiscounted sum of stepwise rewards for the same account from the step when the candidate action takes effect through the end of the period; each stepwise reward retains both the account-level and global feedback terms.

Planner and Optimizer construct leave-one-out advantages using Eq.~\eqref{eq:offline-advantage}, normalizing with an exponential moving average of within-group return standard deviations; candidate groups with overly small return variance receive zero advantages. Inspector uses each candidate branch's return minus that of the branch without a diagnosis. Both share the previous period's feedback, so their difference depends only on next-period outcomes. These advantages stay fixed throughout offline optimization.

Policy optimization uses the sequence-level objective in Eq.~\eqref{eq:offline-loss}. For candidate $g$, let $L_g$ denote the response length, and let $p_{g\ell}(\theta)$ and $p_{g\ell}^b$ denote the conditional probabilities of token $\ell$ under the current and behavior policies, respectively. The length-normalized sequence importance ratio is
\begin{equation}
\varrho_g(\theta)=\exp\left[
\frac{1}{L_g}\sum_{\ell=1}^{L_g}
\log\frac{p_{g\ell}(\theta)}{p_{g\ell}^b}
\right].
\label{eq:offline-ratio}
\end{equation}

The KL penalty uses reference policy $\pi_{\mathrm{ref}}$, with token probability $p_{g\ell}^{\mathrm{ref}}$. Define $\xi_{g\ell}=\log[p_{g\ell}^{\mathrm{ref}}/p_{g\ell}(\theta)]$. The $k3$ estimate averages $\exp(\xi_{g\ell})-\xi_{g\ell}-1$ over valid response tokens. Behavior and reference policies stay fixed throughout offline optimization. Appendix~\ref{sec:training-config} gives the clipping range and KL coefficient.

\subsection{Offline Reinforcement Learning Algorithm}
\label{sec:offline-algorithm}

In Algorithm~\ref{alg:offline}, $\mathcal S_r$ denotes the set of decision snapshots for role $r\in\mathcal R=\{P,O,I\}$, and $\Gamma_r$ maps responses to executable actions or diagnostic guidance. $\mathcal B_r^{\mathrm{valid}}$ stores the inputs, responses, returns, fixed advantages, and behavior policy probabilities and versions of valid candidates, and $\mathcal B$ denotes a training batch.

\begingroup
\SetAlgoSkip{smallskip}
\SetAlFnt{\footnotesize}
\begin{algorithm}[H]
\caption{Offline Reinforcement Learning}
\label{alg:offline}
\KwIn{SFT parameters $\{\theta_r^{\mathrm{SFT}}\}$,  decision snapshots $\{\mathcal S_r\}$,  delivery environment $\mathcal E$.}
\KwOut{Offline parameters $\{\theta_r^{\mathrm{off}}\}$.}
$\theta_r\leftarrow\theta_r^{\mathrm{SFT}}$;  fix $\pi_b^r\leftarrow\pi_{\theta_r}$;  $\mathcal D_r^{\mathrm{off}}\leftarrow\varnothing,\quad\forall r$\;
\For{$r\in\mathcal R,\ \sigma\in\mathcal S_r$}{
  Read $h^r$ from $\sigma$;  $\{z_g^r\}_{g=1}^{G}\sim\pi_b^r(\cdot\mid h^r)$\;
  $u_g^r\leftarrow\Gamma_r(z_g^r,h^r)$;  execute candidates separately from $\sigma$;  compute $\{J_g^r\}_{g=1}^{G}$ as specified in Appendix~\ref{sec:offline-optimization}\;
  \If{$r=I$}{
    Run the branch without a diagnosis;  compute $J_0$ using Eq.~\eqref{eq:role-rewards}\;
  }
  Compute and fix $\{\widehat A_g^r\}$  using Eq.~\eqref{eq:offline-advantage} and its normalization rules\;
  $\mathcal D_r^{\mathrm{off}}\leftarrow\mathcal D_r^{\mathrm{off}}\cup\mathcal B_r^{\mathrm{valid}}$\;
}
\For{$r\in\mathcal R$}{
  Fix $\pi_{\mathrm{ref}}^r$\;
  \For{$\mathcal B\subset\mathcal D_r^{\mathrm{off}}$}{
    Compute $\varrho_g^r$ and $\mathcal L_{\mathrm{off}}^r$  using Eqs.~\eqref{eq:offline-ratio} and~\eqref{eq:offline-loss},  and update $\theta_r$\;
  }
  $\theta_r^{\mathrm{off}}\leftarrow\theta_r$\;
}
\Return{$\{\theta_r^{\mathrm{off}}\}_{r\in\mathcal R}$}\;
\end{algorithm}
\endgroup
\FloatBarrier

\FloatBarrier
\section{Co-Evolution Update Details and Algorithm}
\label{sec:coevo-details}

\setcounter{equation}{0}
\renewcommand{\theequation}{C.\arabic{equation}}
\renewcommand{\theHequation}{C.\arabic{equation}}
\setcounter{algocf}{0}
\renewcommand{\thealgocf}{C.\arabic{algocf}}
\renewcommand{\theHalgocf}{C.\arabic{algocf}}

Co-evolution uses settled feedback from actual execution trajectories. Trajectory $\tau_e$ records inputs, responses, behavior policy versions, generation probabilities, and feedback. Settled records form $\mathcal D_e$ at period end, with role $r$'s records in $\mathcal D_e^r$. The new diagnosis awaits next-period settlement; schedule $k_e$ cycles through $(P,O,I)$.

Each role adds newly settled records to its buffer $\mathcal C^r$ and, when scheduled, selects valid records to form the current round's training set $\mathcal B_e^r$. Used records are removed from the buffer and excluded from later rounds; the remaining records are retained. Updated role policies take effect in the next period.

\subsection{Value Baselines and Advantage Estimation}
\label{sec:gae-details}
\label{sec:value-model}

Each role's independent PrefixValueModel $V_{\phi_r}(h)$ maps the last input token's $D$-dimensional hidden representation before response generation through LayerNorm($D$), Linear($D$, 256), Tanh, and Linear(256, 1) to a scalar return. Detached input features block value-model gradients to the role language model.

Before each training round, the system computes value predictions $V_i^-$, then advantages and return targets. Planner and Inspector use $r_i^r-V_i^-$ as the advantage in Eq.~\eqref{eq:coevo-advantage}, with regression targets $\widehat R_i^r=r_i^r$. Optimizer applies GAE within each account and period, including HOLD decisions.

Fixing one period, let $t_{a,i}$ denote the $i$-th decision step of account $a$, $r_{a,i}^O$ its reward, $\Delta_{a,i}=t_{a,i+1}-t_{a,i}$ the number of environment steps to the next node of the same account, and $b_{a,i}\in\{0,1\}$ the sequence termination indicator. The recursion is

\begin{equation}
\begin{aligned}
\delta_{a,i}&=r_{a,i}^O+\gamma^{\Delta_{a,i}}(1-b_{a,i})V_{a,i+1}^- -V_{a,i}^-,\\
\widehat A_{a,i}^O&=\delta_{a,i}+\gamma^{\Delta_{a,i}}\lambda(1-b_{a,i})\widehat A_{a,i+1}^O,\\
\widehat R_{a,i}^O&=\widehat A_{a,i}^O+V_{a,i}^-.
\end{aligned}
\label{eq:gae-recursion}
\end{equation}

The discount factor is $\gamma=1$ and the GAE parameter is $\lambda=0.95$. Each account's final node in a period has $b_{a,i}=1$ and $\Delta_{a,i}=0$, with zero terminal bootstrap value and boundary advantage. Recursion never crosses accounts or periods, even though rewards include all-account feedback. Advantages and return targets are computed over complete account-period sequences before batching and remain fixed throughout the round.

\subsection{Value Regression and Policy Updates}
\label{sec:value-regression}
\label{sec:sao-details}

The value model performs SmoothL1 regression against fixed, detached return targets $\widehat R_i$, averaging the per-node regression losses over the settled batch $\mathcal B$. Each role's first scheduled round updates only its value model. Later rounds update the value model first, then the role policy using the previously fixed action advantages. The two models maintain separate optimizers and optimizer states.

Policy updates follow Eq.~\eqref{eq:coevo-loss}, with an entire response sharing the same action advantage. Let $p_\ell(\theta)$ and $p_\ell^b$ denote the conditional probabilities of token $\ell$ under the current policy and the behavior policy that collected the response. Their ratio is $\varrho_\ell=p_\ell(\theta)/p_\ell^b$. The selection mask and importance weight are

\begin{equation}
\begin{aligned}
m_\ell&=\mathbf1\!\left[1-\varepsilon^-\leq\varrho_\ell\leq1+\varepsilon^+\right],\\
\bar\varrho_\ell&=\operatorname{sg}\!\left[
\operatorname{clip}(\varrho_\ell,1-\varepsilon^-,1+\varepsilon^+)\right].
\end{aligned}
\label{eq:coevo-filter-weight}
\end{equation}

Here, $\varepsilon^-=\varepsilon^+=0.2$, $\operatorname{sg}$ stops the gradient, and $Z=\sum_{\ell=1}^{L}m_\ell$ is the number of valid tokens.

Each actor update recomputes probability ratios and masks; advantages and importance weights stay fixed during backpropagation. Responses with $Z=0$ are skipped for that update. The $k3$ KL to the behavior policy is used only for monitoring and rollback, not in the actor objective. If KL exceeds 0.01 or the retained-token fraction falls below 0.4, the system restores the parameters and optimizer state from before that update and stops actor updates for the round. Completed value-model updates are retained.

\subsection{Continuous Co-Evolution Algorithm}
\label{sec:coevo-algorithm}

In Algorithm~\ref{alg:coevolution}, let $\Theta_e^r$ denote role $r$'s policy parameters, value parameters, and optimizer states, $\nu_r$ its number of completed training rounds, and $\widetilde\Theta^r$ its state at the end of the current round.

\begin{algorithm}[!htb]
\caption{Continuous Co-Evolution}
\label{alg:coevolution}
\KwIn{Offline parameters $\{\theta_r^{\mathrm{off}}\}$, delivery environment $\mathcal E$, role schedule $\{k_e\}_{e=1}^{E}$.}
\KwOut{Final training states $\{\Theta_{E+1}^r\}$.}
Initialize $\{\Theta_1^r\}$ using $\{\theta_r^{\mathrm{off}}\}$; $\nu_r\leftarrow0$, $\mathcal C^r\leftarrow\varnothing,\ \forall r$; $g_0\leftarrow\varnothing$\;
\For{$e=1,\ldots,E$}{
  Execute and record P/O delivery with $g_{e-1}$ in $\tau_e$; compute P/O rewards using Eqs.~\eqref{eq:coevo-reward-score}--\eqref{eq:role-rewards} and construct $\mathbf y_e$\;
  $z_e^I\sim\pi_{\theta_I}(\cdot\mid h_e^I)$; $g_e\leftarrow\Gamma_I(z_e^I,h_e^I)$\;
  Append the current diagnosis to $\tau_e$ as pending\;
  Initialize $\mathcal D_e$ with settled P/O records; if a previous diagnosis is pending, settle its reward using Eq.~\eqref{eq:role-rewards} and add its record\;
  $\mathcal C^{r'}\leftarrow\mathcal C^{r'}\cup\mathcal D_e^{r'},\quad\forall r'\in\mathcal R$\;
  $r\leftarrow k_e$; form $\mathcal B_e^r$ from valid records in $\mathcal C^r$\;
  Compute and fix $V_i^-,\widehat A_i^r,\widehat R_i^r$ for $\mathcal B_e^r$ using Eq.~\eqref{eq:coevo-advantage} and Appendix~\ref{sec:gae-details}\;
  Divide $\mathcal B_e^r$ into batches; update $\phi_r$ by value regression in Appendix~\ref{sec:value-regression}\;
  \If{$\nu_r>0$}{
    Update $\theta_r$ batch by batch using Eq.~\eqref{eq:coevo-loss} with filtering and rollback in Appendix~\ref{sec:sao-details}\;
  }
  $\mathcal C^r\leftarrow\mathcal C^r\setminus\mathcal B_e^r$\;
  $\nu_r\leftarrow\nu_r+1$; $\Theta_{e+1}^{r}\leftarrow\widetilde\Theta^r$; $\Theta_{e+1}^{r'}\leftarrow\Theta_e^{r'},\ r'\ne r$\;
}
\Return{$\{\Theta_{E+1}^r\}_{r\in\mathcal R}$}\;
\end{algorithm}
\FloatBarrier

\section{Experimental Implementation Settings}
\label{sec:environment-config}

\subsection{Environment Configuration}

Delivery periods use 48 time steps and a traffic generation setting of 500,000 opportunities, following AuctionNet's official defaults~\cite{auctionnet2024}. General is calibrated from the General dataset in AuctionNet's open-source repository: the generation baseline for predicted conversion probabilities (pValue) is 0.005, and budgets and CPA constraints use the full 48-dimensional configurations. Other auction mechanisms remain unchanged.

For each environment configuration, the total traffic volume, initial budgets and CPA constraints, competitor-type composition, and pValue generation baseline remain fixed across periods. Each period uses the episode index to resample predicted conversion probabilities and their variability and within-period traffic curves, and to initialize native competitors' random states, producing different market conditions. Both Sparse and General split episodes 71--320 consecutively into five market sequences, each with 50 delivery periods.

\subsection{Training Configuration}
\label{sec:training-config}

For SFT, each role uses 3,290 demonstration trajectories generated by DeepSeek V4 Flash. Training uses a learning rate of $10^{-4}$, microbatch size 8, gradient accumulation of 2, and maximum total sequence length of 2,560 tokens. Planner, Optimizer, and Inspector are trained for 4, 3, and 3 epochs, respectively.

Offline GSPO uses a learning rate of $10^{-5}$ for 5 epochs, 4 candidates per decision, sequence-ratio clipping to $[0.8,1.2]$, and a KL penalty coefficient of 0.1. Both stages use LoRA with rank 16 and alpha 32.

Co-evolution uses AdamW at actor and value-model learning rates of $10^{-6}$ and $5\times10^{-4}$, respectively. Batches hold valid settled decision nodes for the scheduled role. Rounds with valid nodes update the value model once and the actor at most four times; Appendix~\ref{sec:sao-details} covers the first scheduled round and rollback.

\subsection{Inference and Action Handling}
\label{sec:inference-config}

During continuous delivery, responses are sampled at a temperature of 0.7 with thinking disabled. The context and response limits are 8,192 and 512 tokens, respectively. Top-p is not explicitly specified. Optimizer selects HOLD or a tCPA multiplier of 0.5, 0.6, 0.7, 0.8, 1.0, 1.2, or 1.3.

Foundation models and AdCoEvo use the same role workflow, prompts, output format, and action-parsing rules. We specify the required output format in prompts for foundation models, without enforcing constrained JSON generation. If Planner parsing or format validation fails, the remaining budget is allocated in proportion to the ad units' initial budgets. For Optimizer, parsing failure or an incorrect decision count produces four HOLDs; each invalid multiplier is replaced by HOLD. During evaluation, each invalid response is counted once, regardless of how many errors it contains. Format errors do not trigger resampling.

\subsection{Baseline and Ablation Configurations}
\label{sec:baseline-config}

AdPilot\textsuperscript{*} combines reflection and action generation in a single model call. Its SFT demonstrations are generated by DeepSeek V4 Flash, and its SFT and GRPO training use the same training markets as AdCoEvo.

Rule multiplies an ad unit's budget by 1.1 when its CPA does not exceed the target, and by 0.9 otherwise, then normalizes budgets within the account to preserve its total budget. For tCPA, it uses 0.7 if CPA exceeds 1.10 times the target, 1.3 if CPA is below 0.90 times the target, and HOLD otherwise. No conversions or an exhausted adjustment limit yield HOLD. The rules do not use within-period timing or spending progress to determine when to adjust.

Role-composition ablations omit the excluded roles at inference. Without Optimizer, the execution layer applies HOLD throughout delivery and keeps control tCPA at its initial value; without Planner, each ad unit retains its initial budget limit. Without Inspector, the remaining Planner and Optimizer roles receive the current-period delivery state but no cross-period diagnostic guidance or fixed statistical summaries.

\end{document}